%% file: gradient_diversity.tex
\documentclass[11pt]{article}

\input newcommands

\begin{document}  
\title{Gradient Diversity: \\ a Key Ingredient for Scalable Distributed Learning} 
\author[1]{Dong Yin}
\author[1]{Ashwin Pananjady}
\author[1]{Max Lam}
\author[3]{Dimitris Papailiopoulos}
\author[1]{~~~~~~~~~~~~~~~~~~Kannan Ramchandran}
\author[1,2]{Peter Bartlett}
\affil[1]{Department of Electrical Engineering and Computer Sciences, UC Berkeley}
\affil[2]{Department of Statistics, UC Berkeley}
\affil[3]{Department of Electrical and Computer Engineering, UW Madison}  
\maketitle
 
\input abstract
\input intro

\input related
\input setup

\input convergence

\input stability

\input experiments

\input conclusion

\bibliographystyle{abbrv}
\bibliography{ref}

\appendix 
\section*{Appendix}
\input proof_eg_grad_diversity

\input proof_convergence

\input proof_lower_bound

\input proof_diversity_inducing

\input proof_stability

\end{document}

%% file: newcommands.tex
\usepackage{authblk} 
\usepackage{setspace}
\usepackage{amsfonts,bm,xspace,amsthm,fullpage,amssymb,dsfont}
\usepackage{multirow, algorithm,rotating}
\usepackage{graphicx} 
\usepackage{url}
\usepackage{enumerate}
\usepackage{xcolor,soul}
\usepackage{thm-restate}
\usepackage[small,bf]{caption}
\usepackage{slashbox}
\usepackage{standalone}
\usepackage{verbatim}
\usepackage{setspace}
\usepackage{paralist}
\usepackage{mdwmath}
\usepackage{amssymb}
\usepackage{amsmath}
\usepackage{amsthm}
\usepackage{xfrac}
\usepackage{mathtools}  
\usepackage{chngcntr}
\usepackage{tabulary}
\usepackage{booktabs}
\usepackage{epsfig}
\usepackage{wrapfig}
\usepackage{lipsum}
\usepackage{caption}
\usepackage{subcaption}
\usepackage{multirow}
\usepackage{algcompatible} 
\usepackage{footnote}
\usepackage{paralist}
\usepackage{hyperref}

\newtheorem{theorem}{Theorem}
\newtheorem{definition}{Definition}

\newtheorem{lemma}{Lemma}

\newtheorem{cor}{Corollary}

\newcommand{\sgn}{\operatorname{sign}}
\newcommand{\indi}{\mathds{1}}
\newcommand{\ie}{\emph{i.e.,} }
\newcommand{\eg}{\emph{e.g.,} }

\newcommand{\R}{\mathbb{R}}

\newcommand{\setC}{\mathcal{C}}
\newcommand{\E}{\mathbb{E}}

\newcommand{\vecx}{\mathbf{x}}
\newcommand{\vecw}{\mathbf{w}}
\newcommand{\vecwb}{\mathbf{w}}
\newcommand{\tvecw}{\widetilde{\mathbf{w}}}
\newcommand{\tvecwb}{\widetilde{\mathbf{w}}}

\newcommand{\vecz}{\mathbf{z}}
\newcommand{\tvecz}{\widetilde{\mathbf{z}}}

\newcommand{\matX}{\mathbf{X}}
\newcommand{\gradf}{\nabla f}
\newcommand{\tf}{\widetilde{f}}

\newcommand{\gradF}{\nabla F}

\newcommand{\setS}{\mathcal{S}}

\newcommand{\bigo}{\mathcal{O}}

\newcommand{\W}{\mathcal{W}}
\newcommand{\vect}[1]{\mathbf{#1}}
\newcommand{\mat}[1]{\mathbf{#1}}

\newcommand{\twonm}[1]{\left\|#1\right\|_2}
\newcommand{\twonms}[1]{\|#1\|_2}

\newcommand{\prob}[1]{\mathbb{P}\left\{#1\right\}}
\newcommand{\probs}[1]{\mathbb{P}\{#1\}}
\newcommand{\EXP}[1]{\mathbb{E}\left[#1\right]}
\newcommand{\EE}{\ensuremath{\mathbb{E}}}
\newcommand{\EXPS}[1]{\mathbb{E}[#1]}
\newcommand{\innerp}[2]{\left\langle#1,#2\right\rangle}
\newcommand{\innerps}[2]{\langle#1,#2\rangle}
\newcommand{\abs}[1]{\left|#1\right|}
\newcommand{\abss}[1]{|#1|}
\newcommand{\tsp}{^{\rm T}}


%% file: abstract.tex
\begin{abstract}
It has been experimentally observed that distributed implementations of mini-batch stochastic gradient descent (SGD) algorithms exhibit speedup saturation and decaying generalization ability beyond a particular batch-size.
In this work, we present an analysis hinting that high similarity between concurrently processed gradients may be
a cause of this performance degradation. 
We introduce the notion of {\it gradient diversity} that measures the dissimilarity between concurrent gradient updates, and show its key role in the performance of mini-batch SGD. 
We prove that on  problems with high gradient diversity, mini-batch SGD is amenable to better speedups, while maintaining the  generalization performance of serial (one sample) SGD. 
We further establish lower bounds on convergence where mini-batch SGD slows down beyond a particular batch-size, solely due to the lack of gradient diversity.
We provide experimental evidence indicating the key role of gradient diversity in distributed learning, and discuss how heuristics like dropout, Langevin dynamics, and quantization can improve it.

\end{abstract}

%% file: intro.tex
\section{Introduction}

In recent years, deploying algorithms on distributed computing units has become the {\em de facto}  architectural choice for large-scale machine learning.  Parallel and distributed optimization has gained significant traction  with a large body of recent work establishing near-optimal speedup gains on  both convex and nonconvex objectives \cite{niu2011hogwild,gemulla2011large,dean2012large,yun2013nomad,liu2014asynchronous1,jaggi2014communication,duchi2013estimation, chen2016revisiting}, and several state-of-the-art publicly available (distributed) machine learning frameworks, such as Tensorflow~\cite{abadi2016tensorflow}, MXNet~\cite{chen2015mxnet}, and Caffe2~\cite{chilimbi2014project}, offer distributed implementations of popular learning algorithms.

Mini-batch SGD is the algorithmic cornerstone for several of these distributed frameworks. During a distributed iteration of mini-batch SGD, a master node stores a global model, and $P$ worker nodes compute gradients for $B$ data points, which are randomly sampled from a total of $n$ training data (\ie $B/P$ samples per worker per iteration), with respect to the same global model; the parameter $B$ is commonly referred to as the batch-size.
The master, after receiving these $B$ gradients, applies them to the global model and sends the updated model back to the workers; this is the equivalent of one round of communication. 
The algorithm then continues to its next distributed iteration. 

Unfortunately,  near-optimal scaling for distributed variants of mini-batch SGD is only possible for up to tens of compute nodes. 
Several studies \cite{dean2012large,paleo} indicate that there is a significant gap between ideal and realizable speedups when scaling out to hundreds of compute nodes. 
This commonly observed phenomenon is referred to as  {\it speedup saturation}. 
A key cause of speedup saturation is the communication overheads of mini-batch SGD.

\begin{wrapfigure}{r}{0.3\columnwidth}
\centering
\includegraphics[width=0.3\columnwidth]{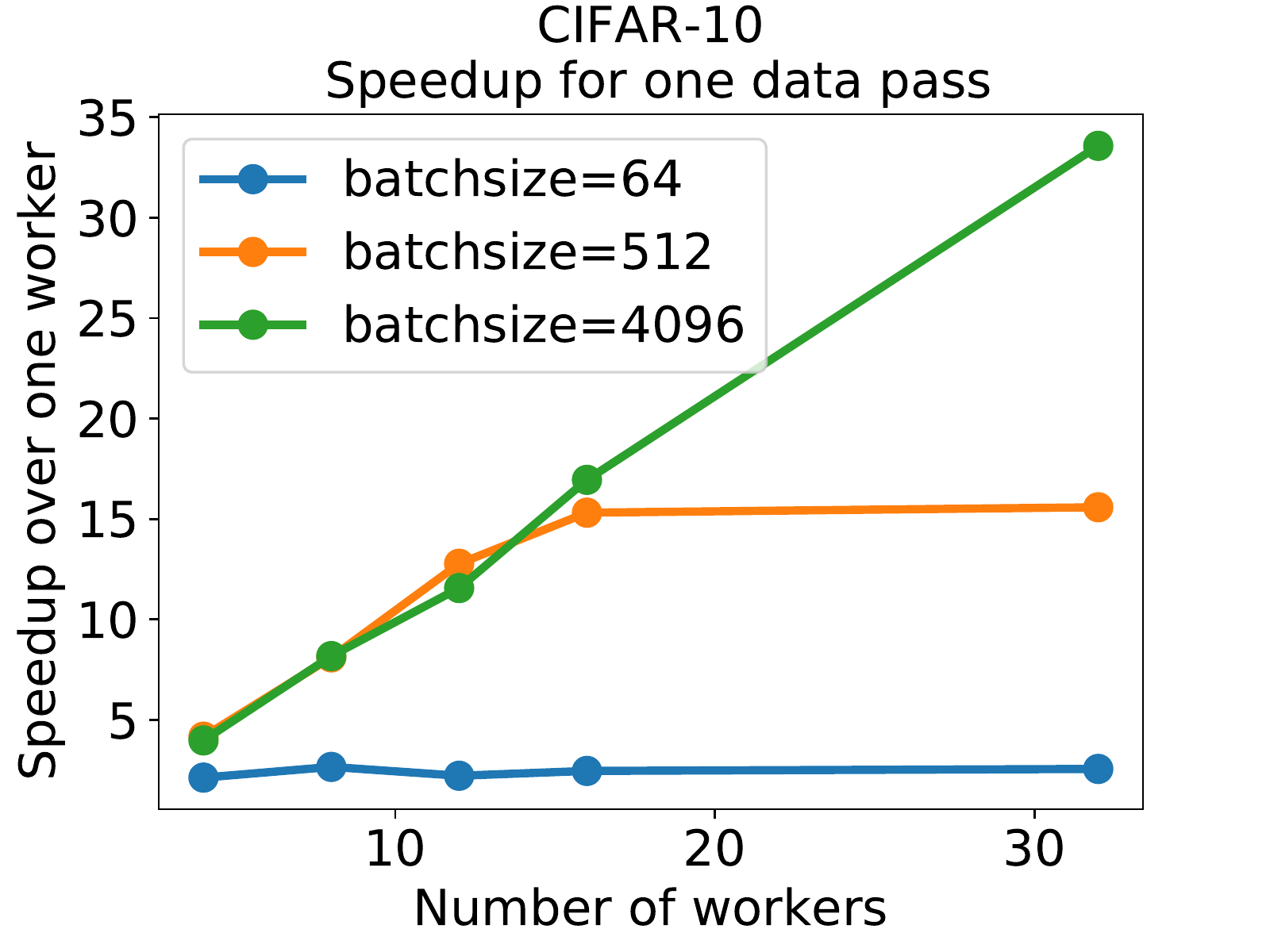}
\caption{\small Speedup gains for a single data pass and various batch-sizes, for a cuda-convnet variant model on CIFAR-10.}
    \centering
    \label{fig:epoch_speedup}
\includegraphics[width=0.3\columnwidth]{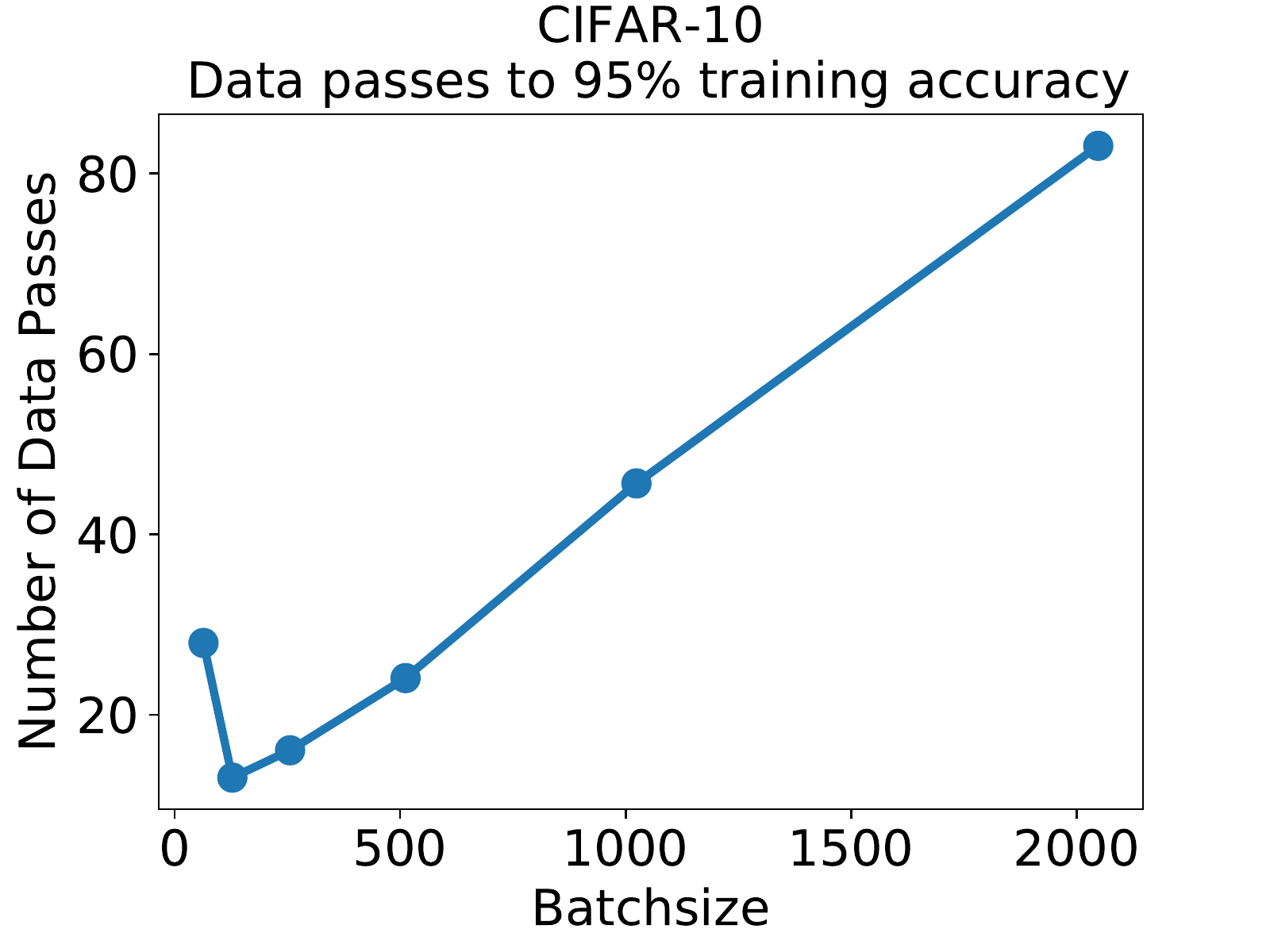}
\caption{\small Number of data passes to reach 95\% accuracy for a cuda-convnet variant  model on CIFAR-10, vs batch-size. Step-sizes are tuned for each batch size to maximize convergence speed.}
\label{fig:batch_vs_acc}
\end{wrapfigure}

Ultimately, the batch-size $B$ controls a crucial performance trade-off between communication costs and convergence speed.
When we use large batch sizes, we observe large speedup gains per pass (\ie per $n$ gradient computations), as shown in Figure~\ref{fig:epoch_speedup}, due to fewer communication rounds. 
However, as shown in Figure~\ref{fig:batch_vs_acc}, to achieve a desired level of accuracy for larger batches, we may need a larger number of passes over the dataset, resulting in {\em overall} slower computation that leads to speedup saturation.
Furthermore, recent work shows that large batch sizes lead to  models that generalize worse~\cite{keskar2016large}.

The key question that motivates our work is the following:
{\it How does the batch-size control the convergence and generalization performance of mini-batch SGD}?


\paragraph{Our Contributions:}

We define the notion of {\it gradient diversity} that measures the dissimilarity between concurrent gradient updates. 
We show that  the convergence  of mini-batch SGD, on both convex and nonconvex objectives, is identical---up to constant factors---to that of serial SGD (\eg $B=1$), if the batch-size is proportional to a bound implied by gradient diversity. We establish that these results are worst-case optimal, \ie there exist convex problems where for larger batches than our prescribed bound, the convergence performance of mini-batch SGD decays. 
To the best of our knowledge this is the first work that presents tight bounds on the batch-size.
Our convergence results are stated for convex, strongly convex, smooth nonconvex, and Polyak-\L{}ojasiewicz functions \cite{karimi2016linear}. Surprisingly, the bound on the optimal batch-size is  identical across all cases.

We note that there has been significant work on the theory of mini-batch algorithms, which we review below. The novelty of our  bound on the optimal batch-size is that it is data-dependent, tight, and essentially identical across convex and nonconvex functions, and in some cases leads to guaranteed uniformly larger batch-sizes compared to prior work. 
More importantly, the bound has an operational meaning and provides insights into algorithmic heuristics like dropout, quantization, and Langevin dynamics, which we show improve gradient diversity.

Following our convergence analysis, we  provide  generalization bounds for mini-batch SGD through the notion of algorithmic stability \cite{bousquet2002stability,hardt2015train,liu2017algorithmic}. 
To the best of our knowledge this is the first work that explores the stability of mini-batch SGD. Through a similar measure of gradient diversity, we establish that as long as the batch-size is below a certain threshold, then mini-batch SGD is as stable as one sample SGD that is analyzed by Hardt et al. \cite{hardt2015train}. 


%% file: related.tex
\section{Related work}
\paragraph{Mini-batch SGD} 
Dekel et al.~\cite{dekel2012optimal} analyze 
mini-batch SGD on non-strongly convex functions
and propose $B=\bigo(\sqrt{T})$ as an optimal choice for batch-size. 
Their result is valid under a particular definition of gap to optimality, and 
does not yield meaningful convergence rates for strongly convex functions, nor does it seem to hold for non-convex setups. 
In contrast, our work provides a general and data-dependent principle for choosing the batch-size via gradient diversity, and it holds without the requirement of convexity.
 Even in the regime where the result in~\cite{dekel2012optimal} is valid, depending on the problem, our result may still provide better bounds on the batch-size than $\bigo(\sqrt{T})$ (\eg in the sparse conflict setting shown in Section~\ref{sec:grad_diversity}). Friedlander and Schmidt~\cite{friedlander2012hybrid} propose an adaptive batch-size scheme that chooses geometrically increasing batch-sizes, and show that this scheme provides weak linear convergence rate for strongly convex functions. 
 With the concept of gradient diversity, our work also implies the similar fact that using a varying batch-size schedule provides a better convergence rate.
 Data-dependent thresholds for batch-size have been developed for some specific problems such as least squares~\cite{jain2016parallelizing} and SVM~\cite{takac2013mini}. These results usually convey a similar message to ours, \ie more diversity among the gradients allows larger batch-size; however, our result holds for a much wider range of problems. In addition to providing theoretical guarantees for the choice of batch-size, De et al. propose an optimization algorithm for choosing the batch-size~\cite{de2016big}. Besides batch-size selection, weighted sampling techniques have also been developed for mini-batch SGD~\cite{needell2016batched,zhang2017stochastic}. In particular, Zhang et al.~\cite{zhang2017stochastic} propose a non-uniform sampling scheme that can increase the chance of getting more diverse data in a batch.

\paragraph{Other mini-batching and distributed optimization algorithms} 
Beyond mini-batch SGD, several other mini-batching algorithms have  been proposed; we survey a non-exhaustive list. 
In ~\cite{li2014efficient,wang2017memory}, mini-batch proximal algorithms are presented that require solving a regularized optimization algorithm on a sampled batch as a subroutine. Although this algorithm allows for choosing a larger batch-size, it also has additional computation and communication cost in distributed settings due to subroutines, and it cannot be trivially applied to the non-convex setting. Accelerated methods in conjunction with mini-batching have been studied in~\cite{cotter2011better}.
Mini-batch SDCA~\cite{shalev2013accelerated,takavc2015distributed} has been proposed for regularized convex problems. The combination of mini-batching and variance reduction has also been studied
in~\cite{nitanda2014stochastic} and mS2GD~\cite{konevcny2016mini}. 
Here, we emphasize that although different mini-batching algorithms can be designed for particular problems and may work better in particular regimes, especially in the convex setting, these algorithms are usually more difficult to implement in distributed learning frameworks like Tensorflow or MXNet, and can introduce additional communication cost.
Other distributed optimization algorithms have also been proposed under different distributed computation frameworks: some algorithms use a one-shot model averaging~\cite{mcdonald2009efficient,zinkevich2010parallelized,zhang2012communication}, and a few other algorithms consider the cases where the workers locally store fractions of the dataset that they do not share it with other workers~\cite{lee2015distributed,shamir2014communication,zhang2015disco,jaggi2014communication}.

\paragraph{Generalization and stability} In their landmark paper~\cite{bousquet2002stability}, Bousquet and Elisseeff show that algorithmic stability implies good generalization. This approach was recently used to establish  generalization bounds for SGD by Hardt et al. \cite{hardt2015train}. Another approach to analyzing the generalization properties of an algorithm is to use the operator view of averaged SGD~\cite{defossez2014constant}. This method was recently extended by Jain et al. \cite{jain2016parallelizing} to the case of random least-squares regression to prove bounds on generalization of mini-batch SGD. In this paper, we extend the analysis of the first method to the mini-batch setting, and show that the generalization is governed by a gradient diversity parameter.

%% file: setup.tex
\section{Problem Setup}

We consider the following general supervised learning setup.
Suppose that $\mathcal{D}$ is an unknown distribution over a sample space $\mathcal{Z}$, and we have access to a sample $\setS = \{\vecz_1,\ldots, \vecz_n\}$ of $n$ data points, that are drawn i.i.d. from $\mathcal{D}$.
Our goal is to find a model $\vecw$ from a model space $\W\subseteq \R^d$ with small \emph{population risk} with respect to a loss function, \ie $R(\vecw)=\E_{\vecz\sim \mathcal{D}}[f(\vecw; \vecz)]$.
Since we do not have access to the population risk, we instead train a model that aims to minimizes the {\it empirical risk}
\begin{equation}\label{eq:erm}
R_{\setS}(\vecw) := \frac{1}{n}\sum_{i=1}^n f(\vecw;\vecz_i).
\end{equation}
For any training algorithm that operates on the empirical risk, there are two important aspects to analyze: the convergence speed to a good model with small empirical risk, and the generalization gap $|R_{\setS}(\vecw)-R(\vecw)|$ that quantifies the performance discrepancy of the model between the empirical and population risks.
For simplicity, we use the notation $f_i(\vecw) := f(\vecw;\vecz_i)$, $F(\vecw) :=R_{\setS}(\vecw)$, and define $\vecw^*\in\arg\min_{\vecw\in\W}F(\vecw)$. 
In this work, we focus on families of differentiable loss functions that satisfy a subset of the following conditions for all parameters $\vecw, \vecw' \in \W$:

\begin{definition}[$\beta$-smooth]\label{def:smooth}	
$F(\vecw) \leq F(\vecw') + \langle \nabla F(\vecw'), \vecw - \vecw' \rangle + \frac{\beta}{2}\|\vecw - \vecw' \|^2_2,$
\end{definition}

\begin{definition}[$\lambda$-strongly convex]
$F(\vecw) \geq F(\vecw') + \langle \nabla F(\vecw'), \vecw - \vecw' \rangle + \frac{\lambda}{2}\|\vecw - \vecw' \|^2_2,$
\end{definition}

\begin{definition}[$\mu$-Polyak-\L{}ojasiewicz (PL)]\label{def:pl}
$\frac{1}{2} \|\nabla F(\vecw) \|_2^2 \geq \mu (F(\vecw) - F(\vecw^*)), \label{eq:pl}$
\end{definition}

\paragraph{Mini-batch SGD}
At each iteration, mini-batch SGD computes $B$ gradients on randomly sampled data points at the most current global model. At the $(k+1)$-th distributed iteration, the mini-batch SGD algorithm is described by

\begin{align}
\vecw_{(k+1)B} = \vecw_{kB} - \gamma \sum_{\ell=kB}^{(k+1)B-1} \gradf_{s_\ell}(\vecw_{kB}), \label{eq:mini-batchupdate}
\end{align}
where each index $s_i$ is drawn uniformly at random from $[n]$ with replacement.
Here, we use $\vecw$ with subscript $kB$ to denote the model we obtain after $k$ distributed iterations, \ie a total number of $kB$ gradient updates. In addition, mini-batch SGD also generally allows varying batch-size $B_1,B_2,\ldots$, and in this case, we will use $\vecw_{N_k}$ to denote the model after $k$ iterations, with $N_k=\sum_{i=1}^kB_i$. Our results also apply to varying step-size, but for simplicity we only state our bounds with constant step-size.
In related studies there is a normalization factor of $1/B$ included in the gradient step, but here we subsume that in the step-size $\gamma$. 
We note that some of our analyses require $\W$ to be a bounded convex subset of $\R^d$, where the projected version of SGD can be used, by making Euclidean projections back to $\W$, \ie
\begin{equation}\label{eq:projected-minibatch}
\vecw_{(k+1)B} = \Pi_\W \left( \vecw_{kB} - \gamma \sum_{\ell=kB}^{(k+1)B-1} \gradf_{s_\ell}(\vecw_{kB}) \right).
\end{equation}
For simplicity, in our main text, we refer to both with/without projection algorithms as ``mini-batch SGD'', but in our Appendix we make the distinction clear, when needed.

%% file: convergence.tex
\section{Gradient Diversity and Convergence}\label{sec:grad_diversity_convergence}
\subsection{Gradient Diversity}\label{sec:grad_diversity}
We introduce the notion of {\it gradient diversity} that quantifies the degree to which individual gradients of the loss functions are different from each other. 

\begin{definition}[Gradient Diversity]
We refer to the following ratio as gradient diversity
\begin{equation}\label{eq:grad_diversity}
\Delta_\setS(\vecw):= \frac{ \sum_{i=1}^n \twonms{\gradf_i(\vecw)}^2 }{\twonms{\sum_{i=1}^n \gradf_i(\vecw)}^2} =\frac{\sum_{i=1}^n \twonms{\gradf_i(\vecw)}^2}{\sum_{i=1}^n \twonms{\gradf_i(\vecw)}^2 + \sum_{i\neq j} \langle\gradf_i(\vecw),\gradf_j(\vecw)\rangle}.
\end{equation}
\end{definition}
We say that $\Delta_\setS(\vecw)$ is a measure of gradient diversity, since it is large when the inner products between the gradients taken with respect to different data points are small. In particular, gradient diversity is large when the gradients are almost orthogonal, or even on opposite directions.
Using this measure of gradient diversity, we define a batch-size bound $B_\setS(\vecw)$ for each data set $\setS$ and each $\vecw\in\W$, as follows. 

\begin{definition}[Batch-size Bound]
$B_\setS(\vecw) := n\cdot \Delta_\setS(\vecw)$.
\end{definition}

As we see in later parts, the batch-size bound $B_\setS(\vecw)$ implied by gradient diversity plays a fundamental role in the batch-size selection during mini-batch SGD.
\paragraph{Examples of gradient diversity} We provide two examples in which we can compute a uniform lower bound 
for all $B_\setS(\vecw)$, $\vecw\in\W$. Notice that these bounds depend on the data set $\setS$,
and are thus \emph{data dependent}.

\noindent\textbf{Example 1.} (Generalized linear function) Suppose that any data point $\vecz$ consists of feature vector $\vecx\in\R^d$ and some label $y\in\R$, and for sample $\setS=\{\vecz_1,\ldots,\vecz_n\}$, the loss function $f(\vecw;\vecz_i)$ can be written as a generalized linear function $ f(\vecw;\vecz_i) = \ell_i(\vecx_i\tsp\vecw)$, where $\ell_i:\R\rightarrow\R$ is a differentiable one-dimensional function, and we do not require the convexity of $\ell_i(\cdot)$. Let $\matX = [\vecx_1~\vecx_2~\cdots~\vecx_n]\tsp \in\R^{n\times d}$ be the feature matrix. We have the following results for $B_\setS(\vecw)$ for generalized linear functions.
\begin{theorem}\label{thm:lbbs}
For generalized linear functions, $\forall~\vecw\in\W$, we have
$$
B_\setS(\vecw) \ge n \frac{\min_{i=1,\ldots,n} \twonms{\vecx_i}^2 }{ \sigma_{\max}^2(\matX)}. 
$$
\end{theorem}

We prove Theorem~\ref{thm:lbbs} in Appendix~\ref{prf:lbbs}. We further instantiate this result in the random feature settings, and provide the following result for features with sub-Gaussian entries.

\begin{cor}\label{cor:subgaussian}
Suppose that $n\ge d$, and $\vecx_i$ has i.i.d. $\sigma$-sub-Gaussian entries with zero mean. Then, there exist universal constants $c_1, c_2, c_3>0$, such that, with probability at least $1-c_2ne^{-c_3d}$, we have $ B_\setS(\vecw) \ge c_1d $ $\forall~\vecw\in\W$. 
\end{cor}
As we can see, as long as we are in the relatively high dimensional regime with $d=\Omega(\log(n))$, with high probability, $B_\setS(\vecw) \ge \bigo(d)$ for all $\vecw\in\W$. We can further improve the probability argument when the magnitude of each entry of the feature matrix is uniformly lower bounded by a positive constant. For example, for Rademacher entries, we have the following result.
\begin{cor}\label{cor:rademacher}
Suppose that $n\ge d$, and the entries of $\vecx_i$ are i.i.d. uniformly distributed in $\{-1,1\}$. Then, there exist universal constants $c_4, c_5, c_6>0$, such that, with probability at least $1-c_5e^{-c_6n}$, we have $B_\setS(\vecw) \ge c_4d$ $\forall~\vecw\in\W$.
\end{cor}

We prove Corollary~\ref{cor:subgaussian} and Corollary~\ref{cor:rademacher} in Appendix~\ref{prf:subgaussian}.

\noindent\textbf{Example 2.} (Loss functions with sparse conflicts) In some applications~\cite{joachims2006training}, the gradient of an individual loss function $\gradf_i(\vecw)$ depends only on a small subset of all the coordinates of $\vecw$ (called the support), and the supports of the gradients have \emph{sparse conflicts}. More specifically, define a graph $G=(V,E)$ with the vertices $V$ representing the $n$ data points, and for $i\neq j$, $(i,j)\in E$ when the supports of $\gradf_i(\vecw)$ and $\gradf_j(\vecw)$ have non-empty overlap. We then have the following result for $B_\setS(\vecw)$.
\begin{theorem}\label{thm:sparse-conflict}
Let $\rho$ be the maximum degree of all the vertices in $G$. Then, we have $\forall~\vecw\in\W$, $B_\setS(\vecw) \ge n/(\rho+1)$. 
\end{theorem}
We prove this result in Appendix~\ref{prf:sparse-conflict}. As we can see, this lower bound can be large when $G$ is sparse, \ie when $\rho$ is small.

\subsection{Convergence Rates}
Our convergence results are consequences of the following lemma, which does not require convexity of the losses and is simple to prove (see Section~\ref{prf:iteration} of the Appendix), and yet captures the effect of mini-batching on an iterate-by-iterate basis. Here, we define $M^2(\vecw) := \frac{1}{n}\sum_{i=1}^n \twonms{\gradf_i(\vecw)}^2$ for any $\vecw\in\W$.

\begin{lemma}\label{lem:iteration}
Let $\vecwb_{kB}$ be a fixed model, and let $\vecw_{(k+1)B}$ denote the model after a mini-batch iteration with batch-size 
$B = \delta \cdot B_\setS(\vecw_{kB})+1$. Then we have:
\begin{equation}\label{eq:cvx_exp_iter3}
\begin{aligned}
&\EXPS{\twonms{\vecwb_{(k+1)B} - \vecw^*}^2\mid\vecwb_{kB}} \\
&\le \twonms{\vecwb_{kB}-\vecw^*}^2 - B\cdot \left(2\gamma \innerp{\gradF(\vecwb_{kB})}{\vecwb_{kB}-\vecw^*}  
-  (1+\delta) \gamma^2 M^2(\vecwb_{kB}) \right).
\end{aligned}
\end{equation}
The inequality changes to equality in the case of minibatch SGD without projections.
\end{lemma}

\paragraph*{Remark.}
For a single iteration, the model trained by serial SGD (\ie $B=1$), in expectation, closes the distance to the optimal by exactly 
$2\gamma \innerp{\gradF(\vecwb_{kB})}{\vecwb_{kB}-\vecw^*}  
- \gamma^2M^2(\vecwb_{kB}) $. Our bound says that, using the \emph{same} step-size\footnote{In fact, our choice of step-size is consistent with many state-of-the-art distributed learning frameworks~\cite{goyal2017accurate}, and we would like to point out that our paper provides theoretical explanation of this choice of step-size.} as SGD (\emph{without normalizing} with a factor of $B$), mini-batch will close that distance to the optimal (or any critical point $\vecw^*$) by approximately $B$ times more, if $B= \bigo(B_\setS(\vecw_{kB}))$. This matches the best that we could have hoped for: mini-batch SGD with batch-size $B$ should be $B$ times faster per iteration than a single iteration of serial SGD.

We now provide convergence rates for strongly convex, convex, smooth nonconvex, and PL functions with constant batch-size. 
For a mini-batch SGD algorithm, define the set $\W_T\subset\W$ as the collection of all possible model parameters that the algorithm can reach during $T/B$ parallel iterations, \ie
$$
\W_T := \{\vecw\in\W~:~\vecw=\vecw_{kB}\text{ for some instance of mini-batch SGD, }k=0,1,\ldots,T/B\}.
$$
Our main message can be summarized as follows:
\begin{theorem}[informal convergence result]\label{thm:convergence_informal}
Let $B\le \delta\cdot B_\setS(\vecw)+1$, $\forall~\vecw\in\W_T$. If serial SGD achieves an $\epsilon$-suboptimal\footnote{Suboptimality is defined differently for different classes of functions.} solution after $T$ gradient updates, then using the same step-size as serial SGD, mini-batch SGD with batch-size $B$ can achieve a $(1+\frac{\delta}{2})\epsilon$-suboptimal solution after the same number of gradient updates (\ie $T/B$ iterations).
\end{theorem}
We can also show that by tuning the step-size by a factor of $1/(1+\delta)$, mini-batch SGD can achieve an $\epsilon$-suboptimal solution using $(1+\delta)T$ gradient updates. 
Therefore, mini-batch SGD does not suffer from convergence speed saturation as long as the batch-size does not exceed the fundamental bound implied by gradient diversity.



Now we provide our convergence results in detail. Define $F^*=\min_{\vecw\in\W}F(\vecw)$, $D_0=\twonms{\vecw_0-\vecw^*}^2$. In all the following results, we assume that $B \le \delta B_\setS(\vecw)+1,~\forall~\vecw\in\W_T$, and $M^2(\vecw) \le M^2,~\forall~\vecw\in\W_T$. The step-sizes in the following results are known to be the order-optimal choices for serial SGD with constant step-size~\cite{bottou2016optimization,ghadimi2016accelerated,karimi2016linear}. 

\begin{theorem}[strongly convex functions]\label{thm:convergence_strong}
Suppose that $F(\vecw)$ is $\lambda$-strongly convex, and use step-size $\gamma = \frac{\epsilon \lambda}{M^2 }$ and batch-size $B\le \frac{1}{2\lambda \gamma}$. Then, after $T\ge \frac{M^2 }{2 \lambda^2 \epsilon }\log(\frac{2D_0}{\epsilon})$ gradient updates, we have
$$
\EXPS{\twonms{\vecwb_T-\vecw^*}^2} \le (1+\frac{\delta}{2})\epsilon.
$$
\end{theorem}

\begin{theorem}[convex functions]\label{thm:convergence_cvx_v1}
Suppose that $F(\vecw)$ is convex, and use step-size $\gamma = \frac{\epsilon}{M^2}$. Then, after $T\ge \frac{M^2D_0}{\epsilon^2}$
gradient updates, we have 
$$
\EXP{F(\frac{B}{T}\sum_{k=0}^{\frac{T}{B}-1}\vecwb_{kB}) - F^*} \le (1+\frac{\delta}{2}) \epsilon.
$$
\end{theorem}

\begin{theorem}[smooth functions]\label{thm:convergence_non_v1}
Suppose that $F(\vecw)$ is $\beta$-smooth, $\W=\R^d$, and use step-size $\gamma = \frac{\epsilon}{\beta M^2}$. Then, after $T \ge \frac{2}{\epsilon^2} M^2 \beta (F(\vecw_0) - F^*)$ gradient updates, we have 
$$
\min_{k=0,\ldots,T/B-1}\EXPS{\twonms{\gradF(\vecwb_{kB})}^2} \le (1+\frac{\delta}{2}) \epsilon.
$$
\end{theorem}

\begin{theorem}[PL functions]\label{thm:convergence_pl}
Suppose that $F(\vecw)$ is $\beta$-smooth, $\mu$-PL, $\W=\R^d$, and use step-size 
$\gamma = \frac{2 \epsilon \mu}{ M^2 \beta}$, and batch-size $B\le \frac{1}{2\gamma \mu}$.
Then, after 
$T \geq \frac{M^2 \beta}{4\mu^2 \epsilon} \log (\frac{2(F(\vecw_0) - F^*)}{\epsilon} )$
gradient updates, we have
$$\EXPS{F(\vecw_T) - F^*} \le (1+\frac{\delta}{2}) \epsilon.
$$
\end{theorem}
We prove Theorems~\ref{thm:convergence_strong},~\ref{thm:convergence_cvx_v1},~\ref{thm:convergence_non_v1}, and~\ref{thm:convergence_pl} in Appendix~\ref{prf:convergence_strong},~\ref{prf:convergence_cvx_v1},~\ref{prf:convergence_non_v1}, and~\ref{prf:convergence_pl}, respectively. 
As mentioned, we can also tune the step-size so that mini-batch SGD can reach $\epsilon$-suboptimality, with a $(1+\delta)$ multiplicative factor on the total number of gradient updates. We present this version of results in Table~\ref{tab:convergence}.

\begin{table}[]
\centering
\begin{tabular}{|c|c|c|}
\hline 
\multicolumn{1}{|l|}{\begin{tabular}[c]{@{}l@{}} Function class\end{tabular}} 
& \begin{tabular}[c]{@{}c@{}}serial SGD\\ step-size $\gamma(\epsilon)$ \end{tabular} 
& \begin{tabular}[c]{@{}c@{}}mini-batch SGD\\ step-size $\gamma(\epsilon)/(1+\delta)$  \end{tabular}  \\ \hline

\begin{tabular}[c]{@{}c@{}}$\lambda$-strongly\\ convex\end{tabular} 
& $\frac{M^2 \log(2D_0 / \epsilon)}{2\lambda^2\epsilon}$ 
& $(1+\delta)\frac{M^2 \log(2D_0 / \epsilon)}{2\lambda^2\epsilon}$  \\ \hline

\multirow{2}{*}{convex} 
& \multirow{2}{*}{$\frac{M^2 D_0}{\epsilon^2}$} 
& \multirow{2}{*}{$(1+\delta)\frac{M^2 D_0}{\epsilon^2}$} \\
  &  &  \\ \hline

\multirow{2}{*}{$\beta$-smooth} 
& \multirow{2}{*}{$\frac{2M^2 \beta \left( F({\bf w}_0) - F^* \right)}{\epsilon^2}$} 
& \multirow{2}{*}{$(1+\delta)\frac{2M^2 \beta \left( F({\bf w}_0) - F^* \right)}{\epsilon^2}$}  \\
&  &    \\ \hline

\begin{tabular}[c]{@{}c@{}}$\beta$-smooth\\ $\mu$-PL \end{tabular} 
& $\frac{M^2 \beta \log \left(2\left(F({\bf w}_0) - F^*\right)/\epsilon\right)}{4\mu^2 \epsilon}$ 
& $(1+\delta)\frac{M^2 \beta \log \left(2 \left( F({\bf w}_0) - F^* \right) / \epsilon \right)}{4\mu^2 \epsilon}$ \\ \hline
\end{tabular}
\caption{Convergence rates of serial SGD and mini-batch SGD with batch-size $B$ for various function classes. For serial SGD, we present the convergence rates that appear in the literature~\cite{bottou2016optimization,ghadimi2016accelerated,karimi2016linear} and we use common choice of step-sizes $\gamma(\epsilon)$, which are also applied in Theorems~\ref{thm:convergence_strong}-\ref{thm:convergence_pl}. For mini-batch SGD, we assume that $B\le \delta\cdot B_\setS(\vecw)+1$, $\forall~\vecw\in\W_T$. The contents of the table show the upper bounds on the total number of gradient updates for each algorithm and each function class to reach $\epsilon$-suboptimality.} \label{tab:convergence}
\end{table}



\subsection{Worst-case Optimality of Fundamental Batch-size}\label{sec:lowerbound}
Interestingly, we can provide a worst-case optimal characterization of convergence rate using $B_\setS(\vecw)$. The following theorem establishes this for a convex problem with varying agnostic batch-sizes\footnote{Here, by saying that the batch-sizes are \emph{agnostic}, we emphasize the fact that the batch-sizes are constants that are picked up without looking at the progress of the algorithm.} $B_k$. Essentially, if we violate the batch bound prescribed above by a factor of $\delta$, then the quality of our model will be penalized by a factor of $\delta$, in terms of accuracy. 

\begin{theorem} \label{thm:lowerboundvarB}
Consider a mini-batch SGD algorithm with $K$ iterations and varying batch-sizes $B_1,B_2,\ldots,B_K$, and let $N_k = \sum_{i=1}^kB_i$. Then, there exists a $\lambda$-strongly convex function $F(\vecw) = \frac{1}{n} \sum_{i=1}^nf_i(\vecw)$ with bounded parameter space $\mathcal{W}$, such that, if $B_k\le \frac{1}{2\lambda\gamma}$ and $B_k \geq \delta \EE[B_\setS (\vecw_{N_{k-1}})] + 1$ $\forall~k=1,\ldots,K$ (where the expectation is taken over the randomness of the mini-batch SGD algorithm), and the total number of gradient updates $T=N_K \ge \frac{c}{\lambda \gamma}$ for some universal constant $c>0$, we have:
$$
\EE [ \| \vecw_{T} - \vecw^* \|_2^2 ] \geq c'(1 + \delta) \frac{\gamma M^2}{\lambda},
$$
where $c'>0$ is a universal constant. More concretely, when running mini-batch SGD with step-size $\gamma = \frac{\epsilon \lambda}{M^2}$
and at least $\bigo(\frac{M^2}{\lambda^2 \epsilon })$ gradient updates, we have
$$
\EE [ \| \vecw_{T} - \vecw^* \|_2^2 ] \geq c'(1 + \delta) \epsilon.
$$
\end{theorem}
We prove Theorem~\ref{thm:lowerboundvarB} in Appendix~\ref{prf:lowerboundvarB}. Although the above bound is only for strongly convex functions, it reveals that there exist regimes beyond which scaling the batch-size beyond our fundamental bound can lead to only worse performance in terms of the accuracy for a given iteration, or the number of iterations needed for a specific accuracy. We would like to note that this result can conceivably be tightened for nonconvex problems, which we suggest as an interesting open problem.

We can also show that, up to a constant factor, the condition $B\le \frac{1}{2\gamma \lambda}$ in Theorem~\ref{thm:convergence_strong}~and~\ref{thm:lowerboundvarB}, is actually necessary for mini-batch SGD to converge when $F(\vecw)$ is strongly convex. We provide the details in Appendix~\ref{prf:strong_smooth_lower}.

\subsection{Diversity-inducing Mechanisms}\label{sec:inducing}
In large scale optimization, a few algorithmic heuristics, such as dropout~\cite{srivastava2014dropout}, stochastic gradient Langevin dynamics (SGLD)~\cite{welling2011bayesian}, and quantization~\cite{tomiokaqsgd}, have been shown to be useful for improving convergence and/or generalization. In this section, we show that these techniques can also increase gradient diversity -- and thus can allow us to use a larger batch-size -- rendering mini-batch SGD more amenable to distributed speedup gains. We note that using these heuristics may also slow down the convergence of mini-batch SGD, since they usually introduce additional noise to the gradients; and there is a trade-off between the speedup gain in distributed system via using large batch-size and the slowdown in convergence rates.

For simplicity, we call any diversity-inducing mechanism a {\sf DIM}. In each iteration, when data point $i$ is sampled, instead of making gradient update with $\gradf_i(\vecw)$, the algorithm updates with a random surrogate vector $\vect{g}_i^{\sf{DIM}}(\vecw)$ by introducing some additional randomness, which is acquired i.i.d. across data points and iterations.

We can thus define the corresponding gradient diversity and batch-size bounds
\begin{equation}\label{eq:def_dim}
\Delta^{\sf{DIM}}_\setS(\vecw):= \frac{ \sum_{i=1}^n \EE\twonms{\vect{g}^{\sf{DIM}}_i(\vecw)}^2 }{\EE\twonms{\sum_{i=1}^n \vect{g}^{\sf{DIM}}_i(\vecw) }^2}, 
\quad B_\setS^{\sf{DIM}}(\vecw):=n\cdot\Delta^{\sf{DIM}}_\setS(\vecw),
\end{equation}
where the expectation is taken over the additional randomness of the mechanism. In the following parts, we first demonstrate various diversity-inducing mechanisms, and then compare $B_\setS^{\sf{DIM}}(\vecw)$ with $B_\setS(\vecw)$.

\paragraph*{Dropout}
We interpret dropout as updating a randomly chosen subset of all the coordinates of the model parameter vector\footnote{We use the interpretation of dropout in~\cite{hardt2015train}, and we note although defined differently, our notion of dropout is of similar spirit to the original dropout~\cite{srivastava2014dropout} and dropconnect~\cite{wan2013regularization} schemes.}. Let $\mat{D}_1,\ldots,\mat{D}_n$ be i.i.d. diagonal matrices with diagonal entries being i.i.d. Bernoulli random variables, and each diagonal entry is $0$ with dropout probability $p\in(0,1)$. When data point $\vecz_i$ is chosen, instead of making gradient update $\gradf_i(\vecw)$, we make update $\vect{g}^{\sf{drop}}_i(\vecw)=\mat{D}_i\gradf_i(\vecw)$.

\paragraph*{Stochastic Gradient Langevin Dynamics}
Adding noise to the gradients, also known as the stochastic gradient Langevin dynamics (SGLD)~\cite{welling2011bayesian} has been shown to improve deep neural network learning~\cite{neelakantan2015adding} and help escape strict saddle points~\cite{ge2015escaping}. SGLD takes the gradient updates: $\vect{g}^{\sf{sgld}}_i(\vecw)=\gradf_i(\vecw)+\vect{\xi}_i$ where $\vect{\xi}_i$, $i=1,\ldots,n$, are independent isotropic Gaussian noise $\mathcal{N}(0,\sigma\mat{I})$. 

\paragraph*{Quantized gradients}
Quantization~\cite{tomiokaqsgd} is a recently proposed technique in deep learning. The quantized version of a vector $\vect{v}$, denoted by $Q(\vect{v})$, is given by
$
[Q(\vect{v})]_\ell = \twonms{\vect{v}} \sgn(v_\ell) \eta_\ell (\vect{v}),
$
where $\sgn(x)$ is the sign of a value $x$, and $\eta_\ell (\vect{v})$s are independent Bernoulli random variables with $\probs{\eta_\ell = 1} = \abss{v_\ell} / \twonms{\vect{v}}$. For quantization, we define $\vect{g}_i^{\sf quant}(\vecw) = Q(\gradf_i(\vecw))$.

We can show that these mechanisms increases gradient diversity, as long as $B_\setS(\vecw)$ is not already large enough. Formally, we have
\begin{theorem}\label{thm:DIM} 
For any $\vecw\in\W$ such that $B_\setS(\vecw) \le n$, we have $B_\setS^{{\sf DIM}}(\vecw) \ge B_\setS(\vecw)$, 
where ${\sf DIM}\in\{\sf{drop, sgld, quant}\}$.
\end{theorem}
We prove Theorem~\ref{thm:DIM} in Appendix~\ref{prf:DIM}. We can also show that if $B_\setS(\vecw)$ is already large enough, \ie $B_\setS(\vecw) > n$, the three mechanisms can still keep the gradient diversity large, \ie $B_\setS^{\sf{DIM}}(\vecw) > n$.

%% file: stability.tex
\section{Differential Gradient Diversity and Stability}

\subsection{Stability and Generalization}
Recall that in supervised learning problems, our goal is to learn a parametric model with small population risk $R(\vecw):=\E_{\vecz\sim \mathcal{D}}[f(\vecw; \vecz)]$. In order to do so, we use empirical risk minimization, and hope to obtain a model that has both small empirical risk and small population risk to avoid overfitting. Formally, let $A$ be a possibly randomized algorithm which maps the training data to the parameter space as $\vecw=A(\setS)$. We define the \emph{expected generalization error} of the algorithm as 
$$
\epsilon_{\text{gen}}(A) :=\abs{\E_{\setS,A}[R_\setS(A(\setS)) - R(A(\setS))]}.
$$
In~\cite{bousquet2002stability}, Bousquet and Ellisseef show that there is a fundamental connection between the generalization error and algorithmic stability. An algorithm is said to be stable if it produces similar models given similar training data. We summarize their result as follows\footnote{This concept of stability is called the average-RO (replacing one) stability in~\cite{shalev2010learnability}.}.
\begin{theorem}\label{thm:generalization}
Let $\setS=(\vecz_1,\ldots, \vecz_n)$ and $\setS^\prime=(\vecz_1^\prime, \ldots, \vecz_n^\prime)$ be two independent random samples from $\mathcal{D}$, and let $\setS^{(i)} = (\vecz_1,\ldots, \vecz_{i-1}, \vecz_i^\prime, \vecz_{i+1}, \ldots, \vecz_n) $ be the sample that is identical to $\setS$ except in the $i$-th data point where we replace $\vecz_i$ with $\vecz_i^\prime$. Then, we have
$$
\E_{\setS,A}[R_\setS(A(\setS)) - R(A(\setS))] = \E_{\setS, \setS^\prime, A} \left[ \frac{1}{n}\sum_{i=1}^n f(A(\setS^{(i)}); \vecz_i^\prime)- \frac{1}{n}\sum_{i=1}^n  f(A(\setS); \vecz_i^\prime) \right].
$$
\end{theorem}
Such a framework was used by Hardt et al.\cite{hardt2015train} to show stability guarantees for serial SGD $(B=1)$, and for Lipschitz and smooth loss functions. Roughly speaking, they show upper bounds $\overline{\gamma}$ on the step-size below which serial SGD is stable. For mini-batch SGD, as our convergence results suggest, in order to gain speed-ups in distributed systems, we would ideally like to operate the mini-batch algorithm using a similar step-size as in serial SGD. We show that the mini-batch algorithm with a similar step-size to SGD is indeed stable, provided that a related notion to gradient diversity is large enough.

\subsection{Differential Gradient Diversity}
The stability of mini-batch SGD is governed by the \emph{differential gradient diversity}, defined as follows. 

\begin{definition}[Differential Gradient Diversity and Batch-size Bound]
For any $\vecw,\vecw'\in\W$, $\vecw\neq\vecw'$, the differential gradient diversity and batch-size bound is given by
\begin{align*}
\overline{\Delta}_\setS(\vecw,\vecw^\prime) := \frac{ \sum_{i=1}^n \twonms{\gradf_i(\vecw) - \gradf_i(\vecw^\prime)}^2 }{\twonms{\sum_{i=1}^n \gradf_i(\vecw) - \gradf_i(\vecw^\prime)}^2},\quad \overline{B}_\setS(\vecw,\vecw^\prime) := n\cdot\overline{\Delta}_\setS(\vecw,\vecw^\prime).
\end{align*}
\end{definition}
Although it is a distinct measure, differential gradient diversity shares similar properties with gradient diversity. For example, the lower bounds for $B_\setS(\vecw)$ in examples 1 and 2 in Section~\ref{sec:grad_diversity} also hold for $ \overline{B}_\setS(\vecw,\vecw^\prime)$, and two mechanisms, dropout and stochastic gradient Langevin dynamics that induce gradient diversity also induce differential gradient diversity, as we note in the Appendix~\ref{sec:example-dgd}.

\subsection{Stability of mini-batch SGD}
We provide the details of our stability result in this section. We make the assumptions that, for each $\vecz\in\mathcal{Z}$, the loss function $f(\vecw;\vecz)$ is convex, $L$-Lipschitz and $\beta$-smooth in $\W$. We choose not to discuss the generalization error for non-convex functions because this, as in \cite{hardt2015train}, requires an significantly small step-size.

Our result is stated informally below, and upper bounds for the generalization error for both convex and strongly convex functions. Here, $\overline{\gamma}$ is the step-size upper bound required to show stability of the serial SGD algorithm, and differently from the convergence results, we treat $\overline{B}_\setS(\vecw,\vecw')$ as a random variable defined by the sample $\setS$.

\begin{theorem}[informal stability result]
Suppose that, with high probability, the batch-size $B\lesssim\overline{B}_\setS(\vecw,\vecw')$ for all $\vecw,\vecw'\in\W$, $\vecw\neq\vecw'$. Then, after the same number of gradient updates, the generalization errors of mini-batch SGD and serial SGD satisfy
$
\epsilon_{\text{gen}}({\sf minibatch~SGD})\lesssim \epsilon_{\text{gen}}({\sf serial~SGD}),
$
and such a guarantee holds for any step-size $\gamma \lesssim \overline{\gamma}$.
\end{theorem}
As one can see, our main message for stability is that, if with high probability, batch-size $B$ is smaller than $\overline{B}_\setS(\vecw,\vecw^\prime)$ for all $\vecw,\vecw'$, mini-batch SGD and serial SGD can be both stable in roughly the \emph{same range} of step-sizes, and the expected generalization error of mini-batch SGD and serial SGD are roughly the \emph{same}.

We now provide our precise theorems bounding the generalization error attained by mini-batch SGD. We use the model parameter obtained in the final iteration as the output of the mini-batch SGD algorithm, \ie $A(\setS)=\vecw_T$.
With the notation in Theorem~\ref{thm:generalization}, we define the following quantity that characterizes the algorithmic \emph{stability} of the learning algorithm given the data points:
\begin{align}
\epsilon_{\text{stab}}(\setS, \setS') = \E_A\left[ \frac{1}{n}\sum_{i=1}^n f(A(\setS^{(i)}); \vecz_i^\prime)- \frac{1}{n}\sum_{i=1}^n  f(A(\setS); \vecz_i^\prime) \right],
\end{align}
where we condition on the data sets $\setS$ and $\setS'$ and take expectation over the randomness of the learning algorithm (mini-batch SGD). Recall from Theorem \ref{thm:generalization} that 
\begin{equation}\label{eq:gen_stab}
\epsilon_{\text{gen}}(A) = \abs{ \E_{\setS, \setS'} \left[ \epsilon_{\text{stab}}(\setS, \setS') \right] } \le \E_{\setS, \setS'} \left[ \abs{\epsilon_{\text{stab}}(\setS, \setS')}  \right].
\end{equation}
We bound $\epsilon_{\text{gen}}(A)$ by first showing a bound on $\epsilon_{\text{stab}}(\setS, \setS')$
that depends on the sample $(\setS, \setS')$, then using equation \eqref{eq:gen_stab} to obtain, as a corollary, results for generalization error.


\paragraph*{Convex Functions}
Our results for convex functions are as follows.
\begin{theorem}[stability of convex functions]\label{thm:stab_cvx}
Fix sample $(\setS, \setS')$. Suppose that for any $\vecz\in\mathcal{Z}$, $f(\vecw;\vecz)$ is convex, $L$-Lipschitz and $\beta$-smooth in $\W$. Provided the step-size and batch-size satisfy
\begin{equation}\label{eq:stab_stepsize}
\gamma \le \frac{2}{\beta \left(1+\frac{1}{n-1}\indi_{B>1}+\frac{B-1}{\overline{B}_\setS(\vecw,\vecw')}\right)},
\end{equation}
for all $\vecw\neq\vecw'$, we have $\abs{ \epsilon_{\text{stab}}(\setS, \setS')} \le 2\gamma L^2 \frac{T}{n}$.
\end{theorem}
Here, $\indi$ denotes the indicator function. Notice that setting $B =1$ recovers the stability result for serial SGD in \cite{hardt2015train} under the same conditions on the step-size, \ie $\gamma\le 2/\beta$, while ensuring a result that holds uniformly for all samples $(\setS, \setS')$. As before, equation \eqref{eq:gen_stab} may be used to directly obtain an upper bound on the generalization error of serial SGD, \ie $\epsilon_{\text{gen}} \le 2\gamma L^2 \frac{T}{n}$. 
For mini-batch SGD, we see that since the sample $(\setS, \setS')$ is random, so is the quantity $\overline{B}_\setS(\vecw,\vecw')$. As a consequence, deriving bounds on the generalization error of the entire algorithm requires understanding the tail behavior of the random variable $\overline{B}_\setS(\vecw,\vecw')$. We provide the following corollary for the generalization error of mini-batch SGD using a tail probability argument.

\begin{cor}[generalization error of convex functions]\label{cor:gen_cvx}
Suppose that for any $\vecz\in\mathcal{Z}$, $f(\vecw;\vecz)$ is convex, $L$-Lipschitz and $\beta$-smooth in $\W$. For a fixed step size $\gamma > 0$, let
\begin{equation}\label{eq:def_eta}
\eta = \prob{\exists~\vecw,\vecw',~\overline{B}_\setS(\vecw,\vecw') < \frac{B-1}{\frac{2}{\gamma \beta} - 1 - \frac{1}{n-1}\indi_{B>1} }},
\end{equation}
where the probability is over the randomness of $\setS$. Then the generalization error of mini-batch SGD satisfies
$$
\epsilon_{\text{gen}} \le 2\gamma L^2 \frac{T}{n}(1-\eta) + 2\gamma L^2 T\eta.
$$
\end{cor}
We prove Theorem~\ref{thm:stab_cvx} and Corollary~\ref{cor:gen_cvx} in Appendix~\ref{prf:stab_cvx}. Notice that when $B=1$, the parameter $\eta=0$, and thus we recover the generalization bound for serial SGD. As we can see, suppose one can find $\overline{B}$ such that 
$\inf_{\vecw\neq\vecw'}\overline{B}_\setS(\vecw,\vecw') \ge \overline{B}$ with high probability, by choosing $B\le 1+\delta \overline{B}$, and
$
\gamma \le \frac{2}{\beta(1+\delta + \frac{1}{n-1})},
$
we can obtain similar generalization error as the serial algorithm without significant change in the step-size range. Equivalently, as long as the batch-size is below the bound implied by differential gradient diversity, we can achieve speedup while keeping the generalization error not significantly affected by mini-batching.

\paragraph*{Strongly Convex Functions} For strongly convex loss functions, we only consider compact and convex parameter space $\W$, and projected mini-batch SGD. Our results take the following form.

\begin{theorem}[stability of strongly convex functions]\label{thm:stab_strong_cvx}
Fix the sample $(\setS, \setS')$. Suppose that for any $\vecz\in\mathcal{Z}$, $f(\vecw;\vecz)$ is $L$-Lipschitz, $\beta$-smooth, and $\lambda$-strongly convex in $\W$, and that $B \le \frac{1}{2\gamma \lambda}$. Provided the step-size and batch-size satisfy
\begin{equation}\label{eq:stab_strong_stepsize}
\gamma \le \frac{2}{(\beta + \lambda) \left(1+\frac{1}{n-1}\indi_{B>1}+\frac{B-1}{\overline{B}_\setS(\vecw,\vecw')}\right)},
\end{equation}
for all $\vecw\neq\vecw'$, we have
$
\abs{ \epsilon_{\text{stab}} (\setS, \setS')} \le \frac{4 L^2}{\lambda n}.
$
\end{theorem}
\begin{cor}[generalization error of strongly convex functions]\label{cor:gen_strong_cvx}
Suppose that for any $\vecz\in\mathcal{Z}$, $f(\vecw;\vecz)$ is $L$-Lipschitz, $\beta$-smooth, and $\lambda$-strongly convex in $\W$, and that $B \le \frac{1}{2\gamma \lambda}$. For a fixed step size $\gamma > 0$, let
\begin{equation}\label{eq:def_eta_strong}
\eta = \prob{\exists~\vecw,\vecw',~\overline{B}_\setS(\vecw,\vecw') < \frac{ B-1 }{ \frac{2}{\gamma (\beta+\lambda)} - 1 - \frac{1}{n-1}\indi_{B>1} }},
\end{equation}
where the probability is over the randomness of $\overline{B}_\setS$. Then the generalization error of mini-batch SGD satisfies 
$$
\epsilon_{\text{gen}} \le \frac{4 L^2}{\lambda n}(1-\eta) + 2\gamma L^2 T\eta.
$$
\end{cor}
We prove Theorem~\ref{thm:stab_strong_cvx} and Corollary~\ref{cor:gen_strong_cvx} in Appendix~\ref{prf:stab_strong_cvx}. We can make similar remarks as the convex case. First, setting $B =1$ recovers the stability result for serial SGD in~\cite{hardt2015train} \ie when $\gamma \le \frac{2}{\beta + \lambda}$, 
$\epsilon_{\text{gen}} \le \frac{4 L^2}{\lambda n}$. Second, if we can find $\overline{B}$ such that $\inf_{\vecw\neq\vecw'}\overline{B}_\setS(\vecw,\vecw') \ge \overline{B}$ with high probability, then we know that as long as we choose $B\le 1+\delta \overline{B}$ and
$
\gamma \le \frac{2}{(\beta+\lambda)\left(1 + \delta +\frac{1}{n-1}\right)},
$
we can achieve a similar generalization error bound as the serial algorithm using the same step size.

\subsection{Examples}
 While in general, the probability parameter $\eta$ may appear to weaken the bound, we can show that there are practical functions of interest for which $\eta$ parameter has fast decay rate in the sample size $n$. For example, we have the following results on the generalization error of mini-batch SGD with generalized linear loss functions and random feature. These results are direct corollaries of the differential gradient diversity bound that we provide in Appendix~\ref{sec:example-dgd}.

\begin{cor}\label{cor:gen_cvx_glm}
Suppose that $f(\vecw;\vecz_i) = \ell_i(\vecz\tsp \vecx_i)$ is $L$-Lipschitz, $\beta$-smooth, and convex in $\W$. In addition, suppose that feature vector $\vecx_i$ has i.i.d. $\sigma$-sub-Gaussian entries. Then there exist universal constants $c_1$, $c_2$, $c_3$, such that when
$$
\gamma \le \frac{2}{\beta(1+\frac{1}{n-1}\indi_{B>1} + c_1\frac{B-1}{d})},
$$
we have
$
\epsilon_{\text{gen}} \le 2\gamma L^2 \frac{T}{n} + c_2\gamma L^2 T ne^{-c_3d}.
$
In addition, if the feature vector $\vecx_i$ has i.i.d. Rademacher entries, the generalization error bound can be improved as
$
\epsilon_{\text{gen}} \le 2\gamma L^2 \frac{T}{n} + c_2\gamma L^2 T e^{-c_3n}.
$
\end{cor}

\begin{cor}\label{cor:gen_strong_glm}
Suppose that $f(\vecw;\vecz_i) = \ell_i(\vecw\tsp \vecx_i)$ is $L$-Lipschitz, $\beta$-smooth, and $\lambda$-strongly convex in $\W$. In addition, suppose that feature vector $\vecx_i$ has i.i.d. $\sigma$-sub-Gaussian entries. Then there exist universal constants $c_1$, $c_2$, $c_3$, such that when
$$
\gamma \le \frac{2}{(\beta+\lambda)(1+\frac{1}{n-1}\indi_{B>1} + c_1\frac{B-1}{d})},
$$
we have$
\epsilon_{\text{gen}} \le \frac{4L^2}{\lambda n} + c_2\gamma L^2 T ne^{-c_3d}$.
In addition, if the feature vector $\vecx_i$ has i.i.d. Rademacher entries, the generalization error bound can be improved as
$
\epsilon_{\text{gen}} \le \frac{4L^2}{\lambda n} + c_2\gamma L^2 T e^{-c_3n}.
$
\end{cor}

As we can see, for generalized linear functions with sub-Gaussian entries, as long as we can in the relatively high dimensional regime ($d=\Omega(\log(n))$ for non-strongly convex functions and $d = \Omega(\log(n) + \log(T))$ for strongly convex functions), mini-batch SGD can achieve generalization error that is of the same order as its serial counterpart without significant change in the range of step-size.

%% file: experiments.tex
\section{Experiments}\label{sec:experiments}
We conduct experiments to justify our theoretical results. Our neural network experiments are all implemented in Tensorflow and run on Amazon EC2 instances g2.2xlarge.

\subsection{Convergence}
\begin{figure}[h]
        \begin{subfigure}[b]{0.33\textwidth}
                \centering
                \includegraphics[width=.99\linewidth]{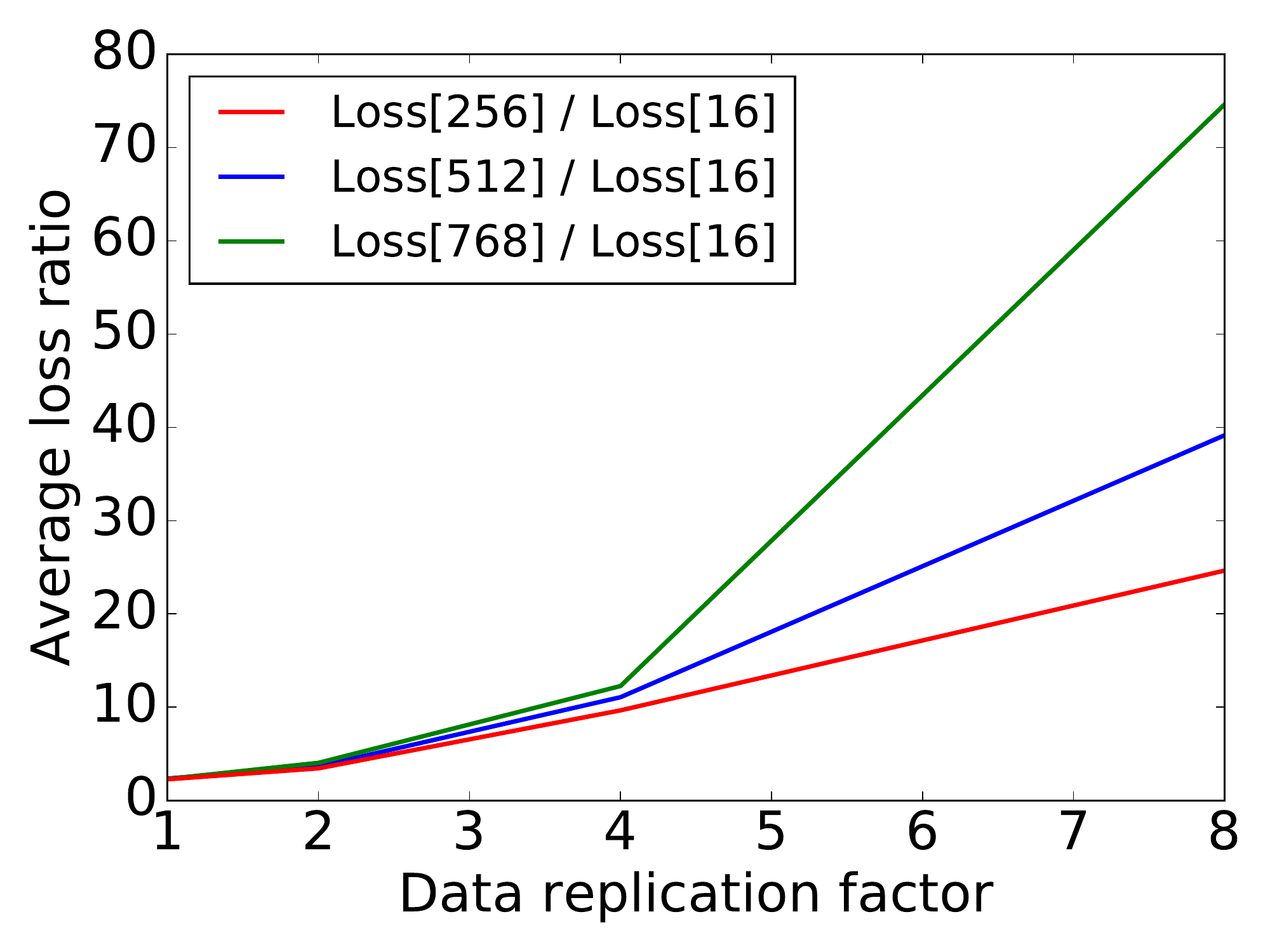}
                \caption{ }
                \label{fig:convexratio}
        \end{subfigure}%
        \begin{subfigure}[b]{0.33\textwidth}
                \centering
                \includegraphics[width=.99\linewidth]{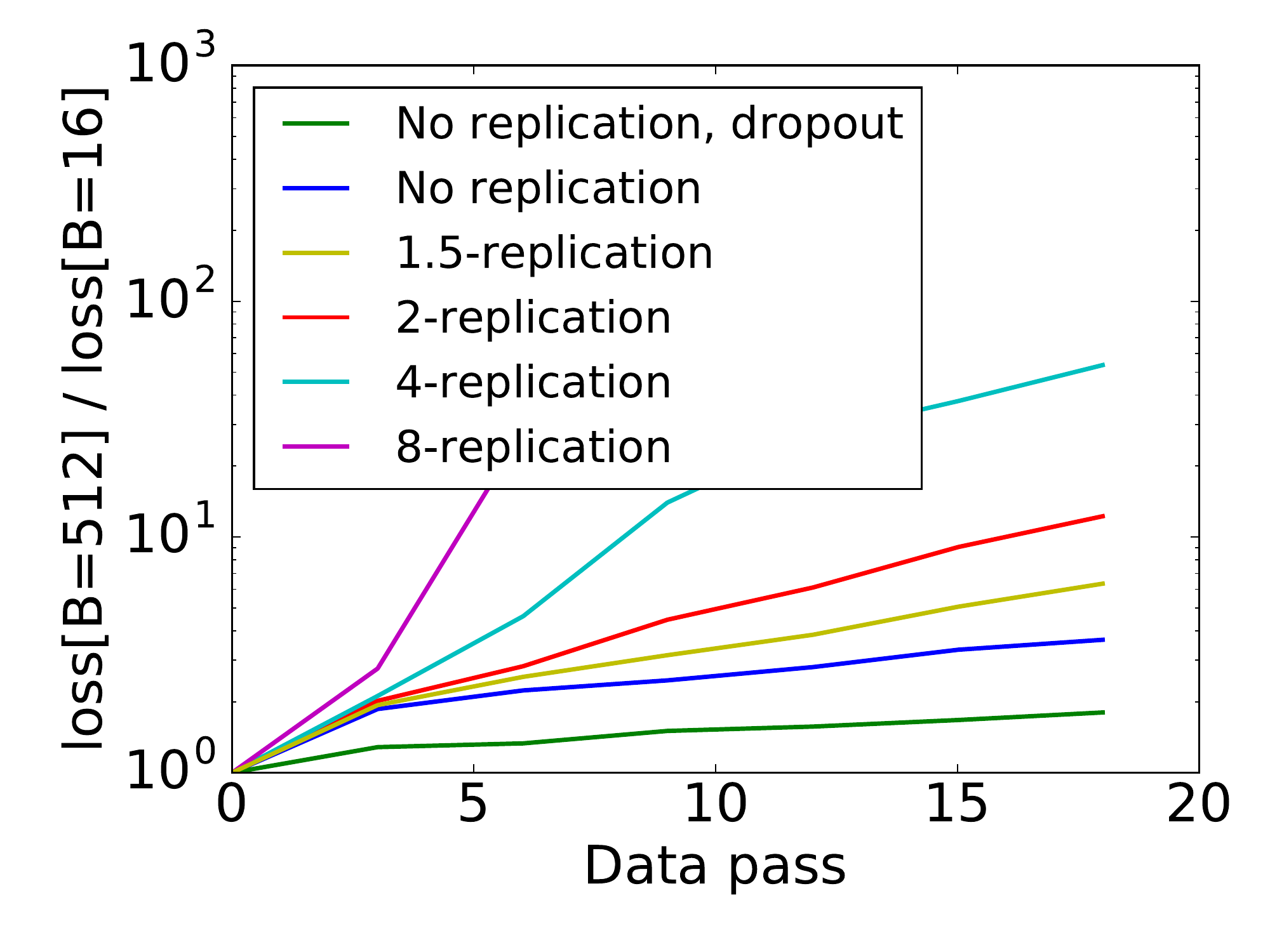}
                \caption{ }
                \label{fig:convnetratio}
        \end{subfigure}%
        \begin{subfigure}[b]{0.33\textwidth}
                \centering
                \includegraphics[width=.99\linewidth]{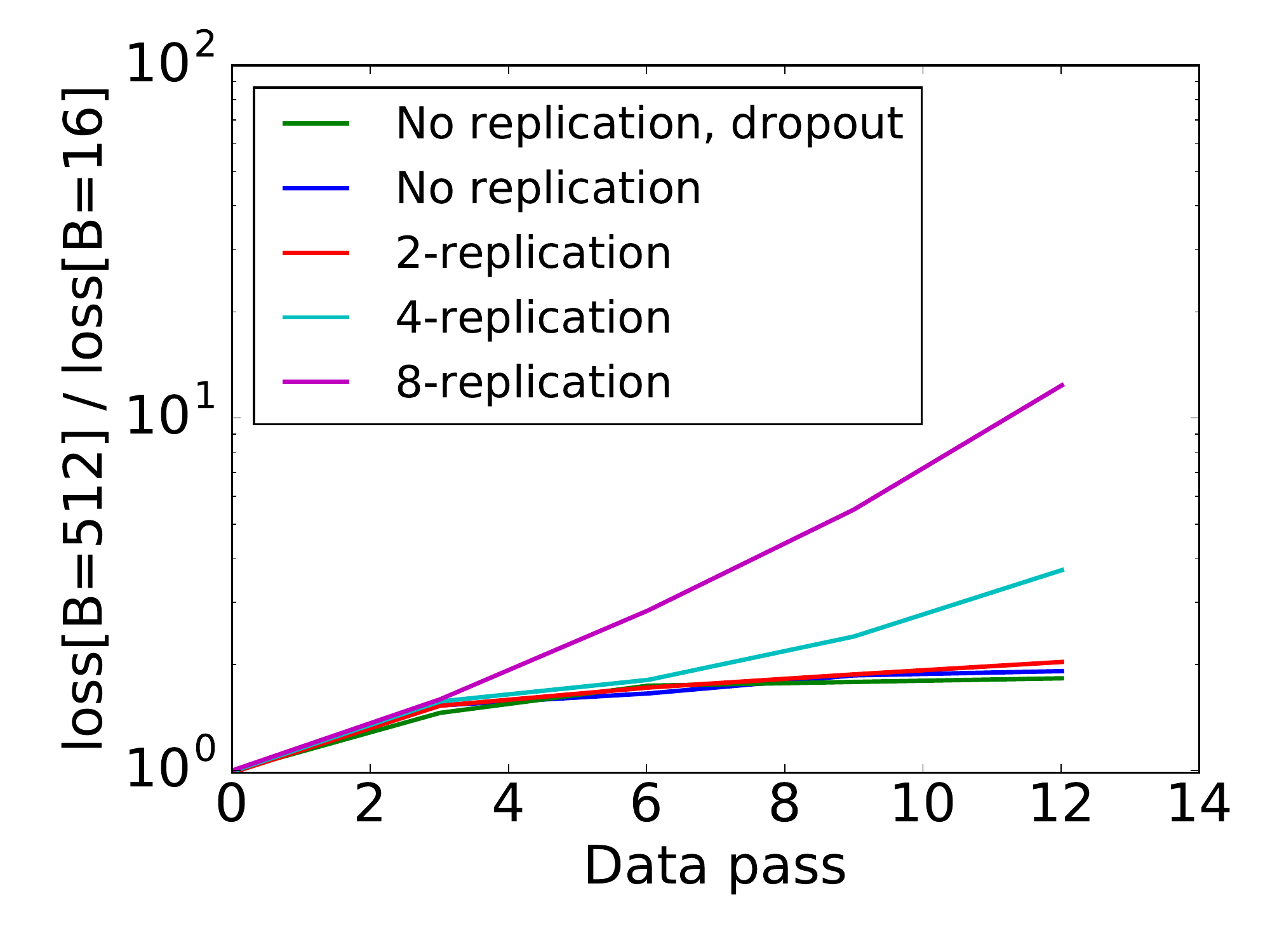}
                \caption{ }
                \label{fig:resnetratio}
        \end{subfigure}
        \caption{Data replication. (a) Logistic regression with two classes of CIFAR-10 (b) Cuda convolutional neural network (c) Residual network. For (a), we plot the average loss ratio during all the iterations of the algorithm, and average over 10 experiments; for (b), (c), we plot the loss ratio as a function of the number of passes over the entire dataset, and average over 3 experiments. Step-sizes are tuned to get fastest convergence for each batch-size.}\label{fig:replication}
\end{figure}

We provide experimental results to justify our theory that higher gradient diversity allows larger batch-size in mini-batch SGD. We conduct the experiments on a logistic regression model and two deep neural networks (a cuda convolutional neural network~\cite{krizhevsky2012imagenet} and a deep residual network~\cite{he2016deep}) with cross-entropy loss running on CIFAR-10 dataset. These results are presented in Figure~\ref{fig:replication}. We use data replication to implicitly construct datasets with different gradient diversity. By replication with a factor $r$ (or $r$-replication), we mean picking a random $1/r$ fraction of the data and replicating it $r$ times. Across all configurations of batch-sizes, we tune the stepsize to maximize convergence. The sample size does not change by data replication, but gradient diversity conceivably gets smaller while we increase $r$. We use the ratio of the loss function for an algorithm instance with large batch-size (\eg $B=512$) to the loss for an algorithm instance with small batch-size (\eg $B=16$) as a metric to measure the negative effect on the convergence rate of using a large batch-size. When this ratio gets larger, the algorithm with the large batch-size is converging slower. We can see from the figures that while we increase $r$, the large batch size instances indeed perform worse, and the large batch instance performs the best when we have dropout, due to its diversity-inducing effect, as discussed in the previous sections. This experiment thus validates our theoretical findings.

\begin{figure}[h]
        \begin{subfigure}[b]{0.33\textwidth}
                \centering
                \includegraphics[width=1.01\linewidth]{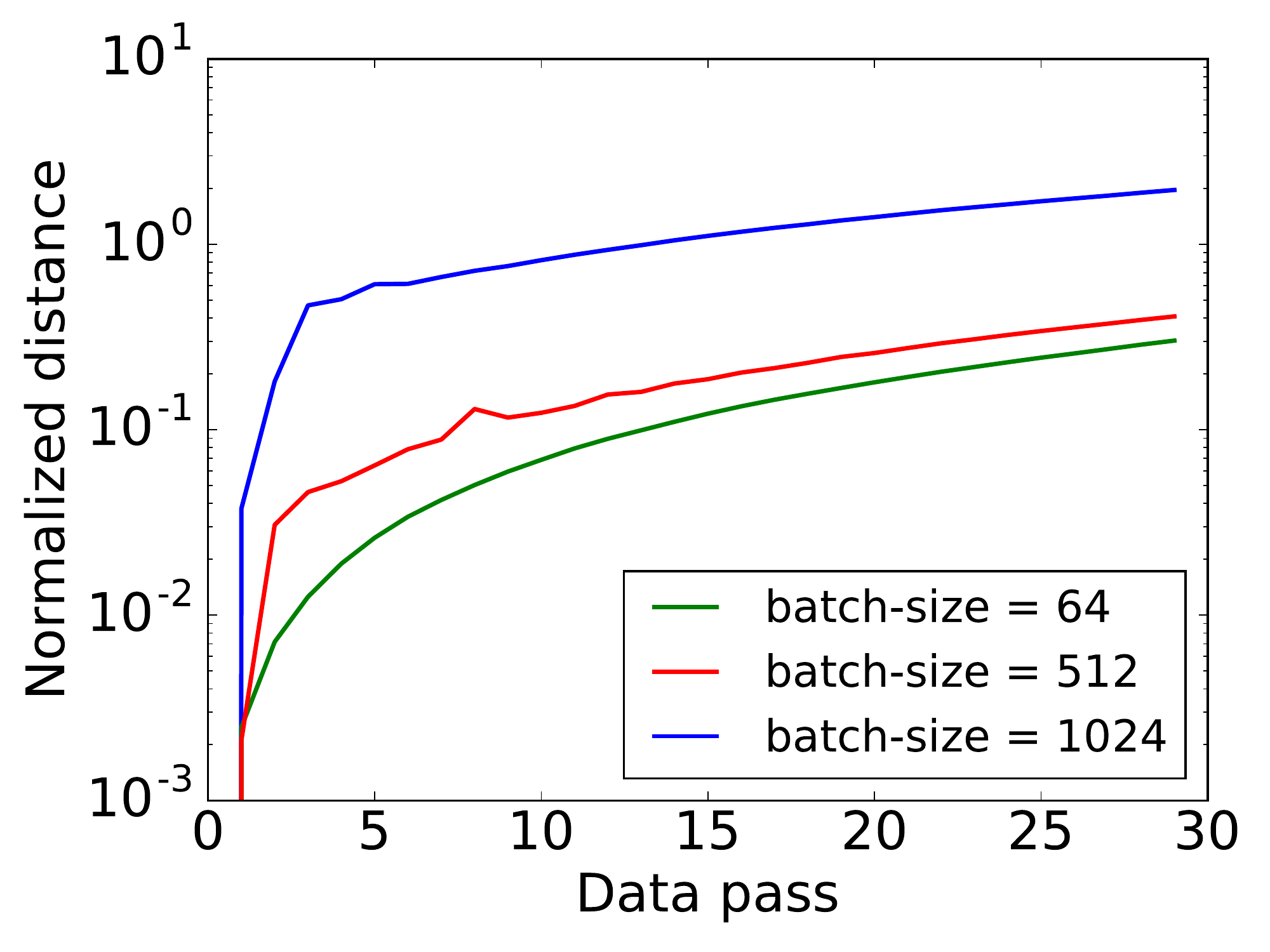}
                \caption{ }
                \label{fig:normdiff}
        \end{subfigure}%
        \begin{subfigure}[b]{0.33\textwidth}
                \centering
                \includegraphics[width=.98\linewidth]{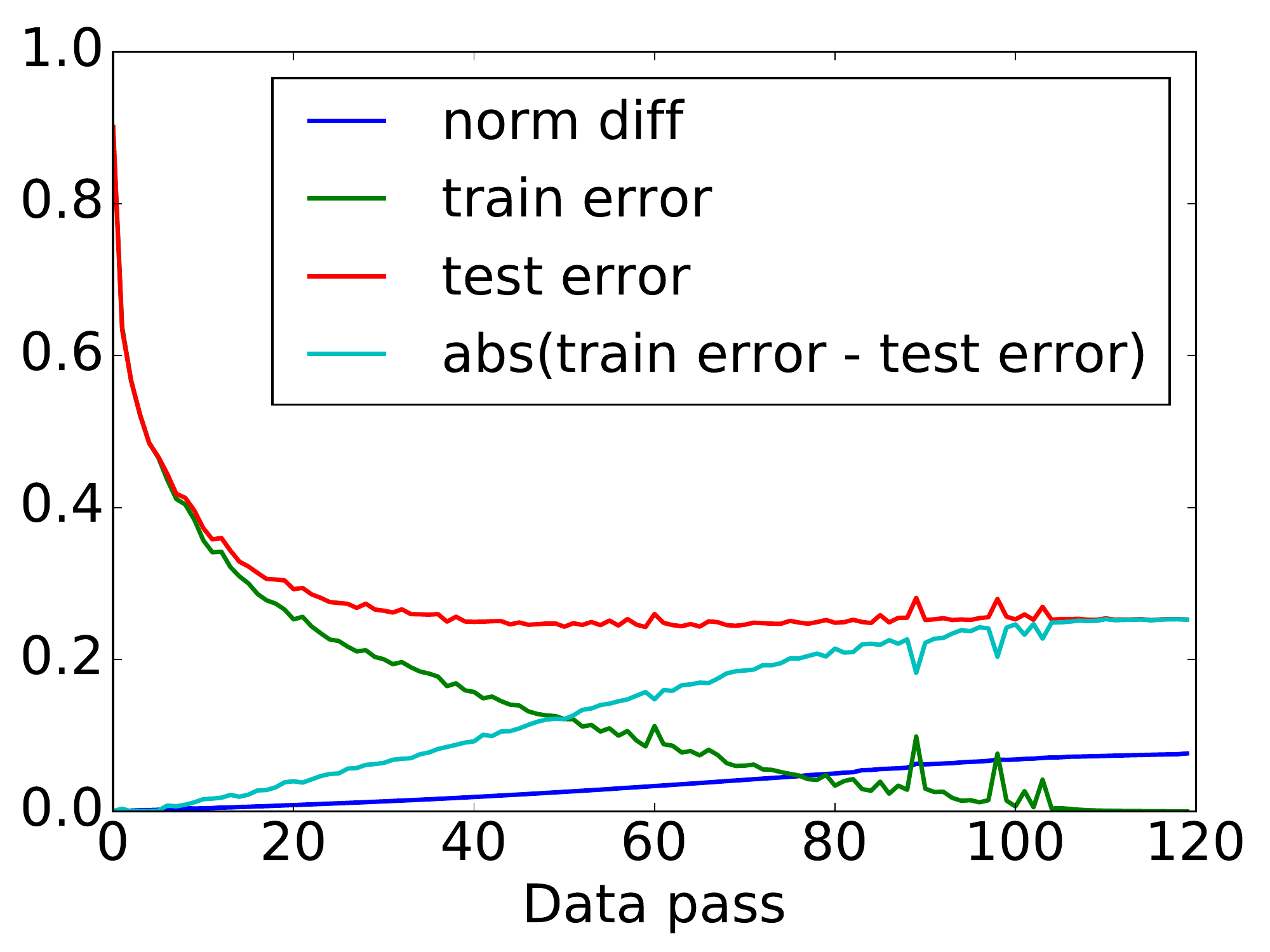}
                \caption{ }
                \label{fig:traintest512}
        \end{subfigure}%
        \begin{subfigure}[b]{0.33\textwidth}
                \centering
                \includegraphics[width=.98\linewidth]{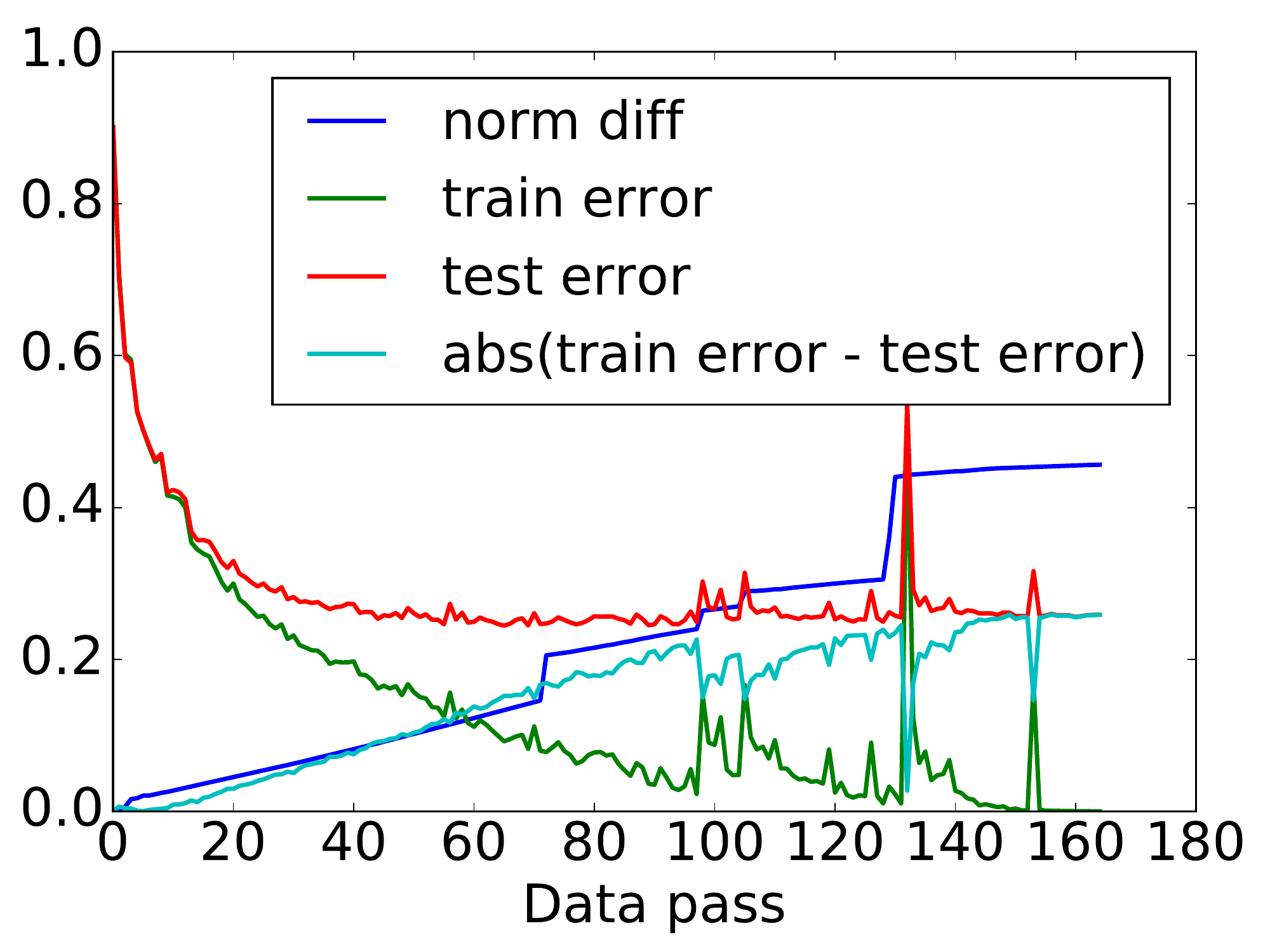}
                \caption{ }
                \label{fig:traintest1024}
        \end{subfigure}
        \caption{Stability. (a) Normalized Euclidean distance vs number of data passes. (b) Generalization behavior of batch-size 512. (c) Generalization behavior of batch-size 1024. Results are averaged over 3 experiments}\label{fig:stability}
\end{figure}

\subsection{Stability}
We also conduct experiments to study the effect of large batch-size on the stability of mini-batch SGD. Our experiments essentially use the same technique as in the study for serial SGD in~\cite{hardt2015train}. Based on the CIFAR-10 dataset, we construct two training datasets which only differ in one data point, and train a cuda convolutional neural network using the same mini-batch SGD algorithm on these two datasets. For different batch-sizes, we test the normalized Euclidean distance $\sqrt{\twonms{\vecw-\vecw'}^2 / (\twonms{\vecw}^2 + \twonms{\vecw'}^2)}$ between the obtained model on the two datasets.
  As shown in Figure~\ref{fig:normdiff}, the normalized distance between the two models becomes larger when we increase the batch-size, which implies that we lose stability by having a large batch-size. We also compare the generalization behavior of mini-batch SGD with $B=512$ and $B=1024$, as shown in Figures~\ref{fig:traintest512} and \ref{fig:traintest1024}. As we can see, for large batch sizes, the models exhibit higher variance in their generalization behavior, and our observation is in agreement with~\cite{keskar2016large}.

%% file: conclusion.tex
\section{Conclusion and Open Problems}
We propose the notion of gradient diversity to measure the dissimilarity between concurrent gradient updates in mini-batch SGD. 
We show that, for both convex and non-convex loss functions, the convergence rate of mini-batch SGD is identical---up to constant factors---to that of serial SGD, provided that the batch-size is at most proportional to a bound implied by gradient diversity. We also develop a corresponding lower bound for the convergence rate of strongly convex objectives. Our results show that on problems with high gradient diversity, the distributed implementation of mini-batch SGD is amenable to better speedups. We also establish similar results for generalization using the notion of differential gradient diversity. Some open problems include finding more mechanisms that improve gradient diversity, and in neural network learning, studying how the network structure, such as width, depth, and activation functions, impacts gradient diversity.

%% file: proof_eg_grad_diversity.tex
\section{Examples of Gradient Diversity}
\subsection{Proof of Theorem~\ref{thm:lbbs}}\label{prf:lbbs}
Let $\ell^\prime(\cdot)$ be the derivative of $\ell(\cdot)$. Since we have
$$
\gradf_i(\vecw) = \ell_i^\prime(\vecx_i\tsp\vecw)\vecx_i,
$$
by letting $a_i := \ell_i^\prime(\vecx_i\tsp\vecw)$ and $\vect{a}=[a_1~\cdots~a_n]\tsp$,  we obtain
$$
B_\setS(\vecw) = \frac{n\sum_{i=1}^n a_i^2\twonms{\vecx_i}^2 }{\twonms{\sum_{i=1}^na_i\vecx_i}^2 } = \frac{n \sum_{i=1}^n a_i^2\twonms{\vecx_i}^2 }{\twonms{\matX\tsp\vect{a}}^2} 
\ge \frac{n \min_{i=1,\ldots,n} \twonms{\vecx_i}^2 \sum_{i=1}^n a_i^2}{ \sigma_{\max}^2(\matX) \twonms{\vect{a}}^2}
\ge \frac{ n \min_{i=1,\ldots,n} \twonms{\vecx_i}^2 }{\sigma_{\max}^2(\matX) },
$$
which completes the proof.

\subsection{ Proof of Corollary~\ref{cor:subgaussian} and Corollary~\ref{cor:rademacher} }\label{prf:subgaussian}
By the concentration results of the maximum singular value of random matrices, we know that when $n\ge d$, there exist universal constants $C_1,C_2,C_3>0$, such that
\begin{equation}\label{eq:bound-max-singular}
\probs{\sigma_{\max}^2(\matX) \le C_1 \sigma^2 n} \ge 1-C_2e^{-C_3n}.
\end{equation}
By the concentration results of sub-Gaussian random variables, we know that there exist universal constants $C_4,C_5>0$ such that
$$
\probs{\twonms{\vecx_i}^2 \ge C_4\sigma^2d} \ge 1-e^{-C_5d},
$$
and then by union bound, we have
\begin{equation}\label{eq:bound-min-norm}
\prob{\min_{i=1,\ldots, n}\twonms{\vecx_i}^2 \ge C_4\sigma^2d} \ge 1-ne^{-C_5d}.
\end{equation}
Then, by combining~\eqref{eq:bound-max-singular} and~\eqref{eq:bound-min-norm} and using union bound, we obtain
$$
\prob{ \frac{ n \min_{i=1,\ldots,n} \twonms{\vecx_i}^2 }{\sigma_{\max}^2(\matX) } \ge \frac{C_4}{C_1}d} \ge 1-C_2e^{-C_3n}-ne^{-C_5d},
$$
which yields the desired result.

Corollary~\ref{cor:rademacher} can be proved using the fact that for Rademacher entries, we have $\twonms{\vecx_i}^2 = d$ with probability one.

\subsection{Proof of Theorem~\ref{thm:sparse-conflict}}\label{prf:sparse-conflict}
We adopt the convention that when $(i,j)\in E$, we also have $(j,i)\in E$. By definition, we have
\begin{align*}
B_\setS(\vecw) &= \frac{n\sum_{i=1}^n \twonms{\gradf_i(\vecw)}^2}{\sum_{i=1}^n \twonms{\gradf_i(\vecw)}^2 + \sum_{i\neq j} \langle\gradf_i(\vecw),\gradf_j(\vecw)\rangle} \\
&= \frac{n\sum_{i=1}^n \twonms{\gradf_i(\vecw)}^2}{\sum_{i=1}^n \twonms{\gradf_i(\vecw)}^2 
+ \sum_{(i,j)\in E} \langle\gradf_i(\vecw),\gradf_j(\vecw)\rangle} \\
&\ge  \frac{n\sum_{i=1}^n \twonms{\gradf_i(\vecw)}^2}{\sum_{i=1}^n \twonms{\gradf_i(\vecw)}^2 
+ \sum_{(i,j)\in E} \frac{1}{2}\twonms{\gradf_i(\vecw)}^2 + \frac{1}{2}\twonms{\gradf_j(\vecw)}^2}.
\end{align*}

Since $\rho$ is the maximum degree of the vertexes in $G$, we know that for each $i\in [n]$, the term $\frac{1}{2}\twonms{\gradf_i(\vecw)}^2$ appears at most $2\rho$ times in the summation $\sum_{(i,j)\in E} \frac{1}{2}\twonms{\gradf_i(\vecw)}^2 + \frac{1}{2}\twonms{\gradf_j(\vecw)}^2$. Therefore, we obtain
$$
\sum_{(i,j)\in E} \frac{1}{2}\twonms{\gradf_i(\vecw)}^2 + \frac{1}{2}\twonms{\gradf_j(\vecw)}^2 \le \rho\sum_{i=1}^n \twonms{\gradf_i(\vecw)}^2,
$$
which completes the proof.

%% file: proof_convergence.tex
\section{Convergence Rates}\label{prf:convergence}
\subsection{Notation}
To assist the demonstration of the proofs of convergence rates, for any $\vecw\in\W$, we define the following two quantities:
$$
M^2(\vecw):=\frac{1}{n}\sum_{i=1}^n\twonms{\gradf_i(\vecw)}^2 \quad\text{and}\quad G(\vecw):=\twonms{\gradF(\vecw)}^2=\twonms{\frac{1}{n}\sum_{i=1}^n\gradf_i(\vecw)}^2
$$
One can check that the batch-size bound obeys $B_\setS(\vecw)=\frac{M^2(\vecw)}{G(\vecw)}$.

\subsection{Proof of Lemma~\ref{lem:iteration}}\label{prf:iteration}
We have
\begin{align*}
\EXPS{\twonms{\vecw_{(k+1)B} - \vecw^* }^2 \mid \vecw_{kB}} =& \EXP{ \twonms{ \vecw_{kB} - \vecw^* - \gamma \sum_{\ell=kB}^{(k+1)B-1} \gradf_{s_\ell}(\vecw_{kB}) }^2  \mid \vecw_{kB}} \\
=&  \twonms{ \vecw_{kB} - \vecw^*}^2 - 2\gamma\sum_{\ell=kB}^{(k+1)B-1}\EXPS{\innerps{\vecw_{kB} - \vecw^*}{\gradf_{s_\ell}(\vecw_{kB})} \mid \vecw_{kB}} \\
&+ \gamma^2 \EXP{\twonms{ \sum_{\ell=kB}^{(k+1)B-1} \gradf_{s_\ell}(\vecw_{kB}) }^2 \mid \vecw_{kB}}.
\end{align*}
Since $s_\ell$'s are sampled i.i.d. uniformly from $[n]$, we know that
\begin{equation}\label{eq:iter-expand}
\begin{aligned}
\EXPS{\twonms{\vecw_{(k+1)B} - \vecw^* }^2 \mid \vecw_{kB}} =&  \twonms{ \vecw_{kB} - \vecw^*}^2 - 2\gamma B \innerps{\vecw_{kB} - \vecw^*}{\gradF(\vecw_{kB})}  \\
&+ \gamma^2 (BM^2(\vecw_{kB}) + B(B-1)G(\vecw_{kB})) \\
=&  \twonms{ \vecw_{kB} - \vecw^*}^2 - 2\gamma B \innerps{\vecw_{kB} - \vecw^*}{\gradF(\vecw_{kB})}  \\
&+ \gamma^2 B \left(1+\frac{B-1}{B_\setS(\vecw_{kB})}\right) M^2(\vecw_{kB}) \\
=&  \twonms{ \vecw_{kB} - \vecw^*}^2 - 2\gamma B \innerps{\vecw_{kB} - \vecw^*}{\gradF(\vecw_{kB})} + \gamma^2 B (1+\delta) M^2(\vecw_{kB}).
\end{aligned}
\end{equation}
We also mention here that this result becomes inequality for the projected mini-batch SGD algorithm, since Euclidean projection onto a convex set is non-expansive.

\subsection{Proof of Theorem~\ref{thm:convergence_strong}}\label{prf:convergence_strong}
According to Lemma~\ref{lem:iteration}, we have
$$
\EXPS{\twonms{\vecwb_{(k+1)B} - \vecw^*}^2\mid\vecwb_{kB}} \le \twonms{\vecwb_{kB}-\vecw^*}^2 - 2\gamma B\innerp{\gradF(\vecwb_{kB})}{\vecwb_{kB}-\vecw^*} + (1+\delta)\gamma^2 BM^2(\vecwb_{kB}).
$$
By strong convexity of $F(\vecw)$, we have
$$
\innerp{\gradF(\vecwb_{kB})}{\vecwb_{kB}-\vecw^*} \ge \lambda \twonms{\vecwb_{kB}-\vecw^*}^2,
$$
which yields
\begin{equation}\label{eq:strong_iter}
\EXPS{\twonms{\vecwb_{(k+1)B} - \vecw^*}^2\mid\vecwb_{kB}} \le (1- 2\gamma \lambda B)\twonms{\vecwb_{kB}-\vecw^*}^2 + (1+\delta)\gamma^2 BM^2 (\vecwb_{kB}).
\end{equation}

Then, by taking expectations over the randomness of the whole algorithm on both sizes of~\eqref{eq:strong_iter}, we obtain
$$
\EXPS{\twonms{\vecwb_{(k+1)B} - \vecw^*}^2 } \le (1 - 2\gamma \lambda B)\EXPS{\twonms{\vecwb_{kB}-\vecw^*}^2} + (1+\delta)\gamma^2 B M^2.
$$
Then if $B\le \frac{1}{2 \gamma \lambda}$, we obtain
$$
\EXPS{\twonms{\vecwb_{T} - \vecw^*}^2 } \le (1- 2\gamma \lambda B)^{T/B}\twonms{\vecwb_{0} - \vecw^*}^2+ (1+\delta)\frac{\gamma M^2 }{2\lambda}.
$$
Using the fact that $1-x\le e^{-x}$ for any $x\ge 0$, we get
$$
\EXPS{\twonms{\vecwb_{T} - \vecw^*}^2 } \le e^{-2\gamma \lambda T}D_0 + (1+\delta)\frac{\gamma M^2}{2\lambda}.
$$
We complete the proof by taking $\gamma = \frac{\epsilon \lambda}{M^2 }$.

\subsection{Proof of Theorem~\ref{thm:convergence_cvx_v1}}\label{prf:convergence_cvx_v1}
According to Lemma~\ref{lem:iteration}, for every $k=0,1,\ldots, \frac{T}{B}-1$, we have
$$
\EXPS{\twonms{\vecwb_{(k+1)B} - \vecw^*}^2 \mid \vecwb_{kB}} \le \twonms{\vecwb_{kB}-\vecw^*}^2 - 2\gamma B\innerp{\gradF(\vecwb_{kB})}{\vecwb_{kB}-\vecw^*}
 +(1+\delta)\gamma^2 B M^2.
$$
Then, we take expectation over all the randomness of the algorithm. Let $D_{kB} = \EXPS{\twonms{\vecwb_{kB}-\vecw^*}^2}$. We have
\begin{equation}\label{eq:iter2}
\EXPS{\innerps{\gradF(\vecwb_{kB})}{\vecwb_{kB} - \vecw^*}} \le \frac{1}{2\gamma B}(D_{kB}-D_{(k+1)B}) + (1+\delta)\frac{\gamma}{2}M^2 .
\end{equation}
We use~\eqref{eq:iter2} to prove the convergence rate. We have by convexity
\begin{align*}
\EXP{F\left(\frac{B}{T}\sum_{k=0}^{\frac{T}{B}-1}\vecwb_{kB}\right) - F(\vecw^*)} &\le \EXP{\frac{B}{T}\sum_{k=0}^{\frac{T}{B}-1}F(\vecwb_{kB}) - F(\vecw^*)} \\
&= \frac{B}{T}\sum_{t=0}^{\frac{T}{B}-1}\EXPS{F(\vecwb_{kB}) - F(\vecw^*)} \\
&\le \frac{B}{T}\sum_{t=0}^{\frac{T}{B}-1} \EXPS{\innerps{\gradF(\vecwb_{kB})}{\vecwb_{kB} - \vecw^*}} \\
&\le \frac{D_0 }{2\gamma T} + (1+\delta)\frac{\gamma M^2 }{2},
\end{align*}
where the last inequality is obtained by taking a summation of (\ref{eq:iter2}) over $k=0,1,\ldots, \frac{T}{B}-1$. Then, we can derive the results by replacing $\gamma$ and $T$ with the particular choices.

\subsection{Proof of Theorem~\ref{thm:convergence_non_v1}}\label{prf:convergence_non_v1}
Recall that we have the iteration
$
\vecwb_{(k+1)B} = \vecwb_{kB} - \gamma\sum_{t=kB}^{(k+1)B-1}\gradf_{s_t}(\vecwb_{kB}).
$
Since $F(\vecw)$ has $\beta$-Lipschitz gradients, we have
$$
F(\vecwb_{(k+1)B}) \le F(\vecwb_{kB}) + \innerps{\gradF(\vecwb_{kB})}{\vecwb_{(k+1)B}-\vecwb_{kB}} + \frac{\beta}{2}\twonms{\vecwb_{(k+1)B}-\vecwb_{kB}}^2.
$$
Then, we obtain
$$
\innerp{\gradF(\vecwb_{kB})}{\gamma\sum_{t=kB}^{(k+1)B-1}\gradf_{s_t}(\vecwb_{kB})} \le F(\vecwb_{kB}) - F(\vecwb_{(k+1)B}) + \frac{\beta}{2}\twonm{\gamma\sum_{t=kB}^{(k+1)B-1}\gradf_{s_t}(\vecwb_{kB})}^2.
$$
Now we take expectation on both sides. By iterative expectation, we know that for any $t\ge kB$,
$$
\EXPS{\innerps{\gradF(\vecwb_{kB})}{\gradf_{s_t}(\vecwb_{kB})}} = \EXPS{\twonms{\gradF(\vecwb_{kB})}^2}.
$$
We also have
$$
\EXP{\twonm{\sum_{t=kB}^{(k+1)B-1}\gradf_{s_t}(\vecwb_{kB})}^2} = \EXPS{BM^2 (\vecwb_{kB}) + B(B-1)G(\vecwb_{kB})} \le B (1+\delta) M^2 .
$$
Consequently,
\begin{equation}\label{eq:non_cvx_iter}
\gamma B \EXPS{\twonms{\gradF(\vecwb_{kB})}^2} \le \EXPS{F(\vecwb_{kB})} - \EXPS{F(\vecwb_{(k+1)B})} + \frac{\beta}{2}\gamma^2 B(1+\delta)M^2 .
\end{equation}
Summing up equation~\eqref{eq:non_cvx_iter} for $k=0,\ldots,T/B-1$ yields 
$$
\gamma B \sum_{k=0}^{T/B-1}\EXPS{\twonms{\gradF(\vecwb_{kB})}^2} \le F(\vecwb_0) - F^* + \frac{\beta}{2}\gamma^2 T(1+\delta)M^2,
$$
which simplifies to
$$
\min_{k=0,\ldots,T/B-1}\EXPS{\twonms{\gradF(\vecwb_{kB})}^2} \le \frac{F(\vecwb_0) - F^*}{\gamma T} + \frac{\beta}{2}\gamma (1+\delta)M^2 .
$$
We can then derive the results by replacing $\gamma$ and $T$ with the particular choices.

\subsection{Proof of Theorem~\ref{thm:convergence_pl}}\label{prf:convergence_pl}
Substituting $\vecw = {\vecw}_{(k+1)B}$ and $\vecw' = \vecw_{kB}$ in the condition for $\beta$-smoothness in Definition~\ref{def:smooth}, we obtain
\begin{align}
F({\vecw}_{(k+1)B}) \leq F(\vecw_{kB}) - \gamma \innerp{\nabla F(\vecw_{kB})}{ \sum_{t=kB}^{(k+1)B-1} \nabla f_{s_t} (\vecw_{kB})} + \frac{\beta \gamma^2}{2} \twonm{ \sum_{t=kB}^{(k+1)B-1} \nabla f_{s_t} (\vecw_{kB}) }^2.
\end{align}
Condition on $\vecw_{kB}$ and take expectations over the choice of $s_t$, $t=kB,\ldots, (k+1)B-1$. We obtain
\begin{align}
\E[F({\vecw}_{(k+1)B})\mid \vecw_{kB}] \leq F(\vecw_{kB}) - \gamma B \| \nabla F(\vecw_{kB})\|_2^2 + \frac{\beta \gamma^2}{2} \left( B M^2(\vecw_{kB}) + B(B-1) G(\vecw_{kB})\right). \label{eq:pl1}
\end{align}
Then, we take expectation over all the randomness of the algorithm. Using the PL condition in Definition~\ref{def:pl} and the fact that $B \le 1+ \delta B_\setS(\vecw)$ for all $\vecw\in\W_T$, we write
\begin{align}\label{eq:pl_form1}
\EE \left[ F({\vecw}_{(k+1)B}) - F^* \right] \leq (1 - 2 \gamma \mu B) \EE \left[ F(\vecw_{kB}) - F^* \right] + (1+\delta)\frac{\beta B \gamma^2 M^2}{2}.
\end{align}
Then, if $B \le \frac{1}{2\gamma\mu}$, we have
$$
\EE \left[ F({\vecw}_T) - F^* \right] \leq (1 - 2 \gamma \mu B)^{T/B} ( F(\vecw_{0}) - F^* ) + (1+\delta)\frac{\beta \gamma M^2}{4\mu}.
$$
Using the fact that $1-x\le e^{-x}$ for any $x\ge 0$, and choosing $\gamma = \frac{2 \epsilon \mu}{ M^2 \beta}$, we get the desired result.

%% file: proof_lower_bound.tex
\section{Lower Bound}
\subsection{Proof of Theorem~\ref{thm:lowerboundvarB}}\label{prf:lowerboundvarB}
We set $f_i(\vecw) = \frac{\lambda}{2}\twonms{\vecw - \vecx_i}^2$, and thus $F(\vecw) = \frac{1}{n}\sum_{i=1}^n\frac{\lambda}{2}\twonms{\vecw- \vecx_i}^2$. We choose
$\W = \{\vecw : \twonms{\vecw} \le 1\}$, and $\vecx_i$'s such that $\twonms{\vecx_i}=1$ for all $i=1,\ldots,n$, and $\sum_{i=1}^n \vecx_i = \vect{0}$.

One can check that $\gradf_i(\vecw) = \lambda(\vecw - \vecx_i)$, $\gradF(\vecw) = \lambda\vecw$, and 
$$
M^2(\vecw) = \frac{1}{n}\sum_{i=1}^n \twonms{\gradf_i(\vecw)}^2= \frac{1}{n} \sum_{i=1}^n \lambda^2 \twonms{\vecw-\vecx_i}^2 = \frac{1}{n} \sum_{i=1}^n \lambda^2 (\twonms{\vecw}^2+\twonms{\vecx_i}^2 ).
$$
Since 
$
M^2(\vecw) = \frac{1}{n} \sum_{i=1}^n \lambda^2 (\twonms{\vecw}^2+\twonms{\vecx_i}^2 ) \in [\lambda^2, 2\lambda^2]
$
for all $\vecw \in \W$, we know that we have $M^2(\vecw) \ge \frac{1}{2}M^2$ for all $\vecw \in \W$. 

Since $\W$ is a bounded set, the projection step has to be taken in order to guarantee that $\vecw_{N_k}\in\W$. However, one can show that, if the initial guess $\vecw_0$ is in the convex hull of $\vecx_1,\ldots,\vecx_n$ (denoted by $\setC\subset\W$), then, without using projection, the obtained model parameter $\vecw_{N_k}$ always stays inside $\setC$. More specifically, we have the following result.
\begin{lemma}\label{lem:convex-hull}
Suppose that $B_k \le \frac{1}{\lambda \gamma}$ for all $k = 1,\ldots, K$, and $\vecw_0\in\setC$. Then, without using projection, $\vecw_{N_{k}}\in\setC$ for all $k$.
\end{lemma}
\begin{proof}
We prove this result using induction. Suppose that $\vecw_{N_{k-1}}\in\setC$. Then, we have
\begin{align*}
\vecw_{N_k} = &\vecw_{N_{k-1}} - \gamma \sum_{\ell = N_{k-1}}^{N_k-1} \gradf_{s_\ell}(\vecw_{N_{k-1}}) =\vecw_{N_{k-1}} - \gamma \sum_{\ell = N_{k-1}}^{N_k-1} \lambda(\vecw_{N_{k-1}} - \vecx_{s_\ell}) \\
=&(1-\gamma \lambda B_k) \vecw_{N_{k-1}} + \gamma \lambda B_k\left(\frac{1}{B_k}\sum_{\ell = N_{k-1}}^{N_k-1} \vecx_{s_\ell} \right).
\end{align*}
Since $\vecw_{N_{k-1}}, \frac{1}{B_k}\sum_{\ell = N_{k-1}}^{N_k-1} \vecx_{s_\ell}\in\setC$, we prove Lemma~\ref{lem:convex-hull}.
\end{proof}
From now on we assume $\vecw_0\in\setC$ and do not consider projection. According to~\eqref{eq:iter-expand} in the proof of Lemma~\ref{lem:iteration}, we have\footnote{We still keep $\vecw^*$ although $\vecw^*=\vect{0}$.}
\begin{align*}
\EXPS{\twonms{\vecw_{N_k} - \vecw^* }^2 \mid \vecw_{N_{k-1}}} = & \twonms{ \vecw_{N_{k-1}} - \vecw^*}^2 - 2\gamma B_k \innerps{\vecw_{N_{k-1}} - \vecw^*}{\gradF(\vecw_{N_{k-1}})}  \\
&+ \gamma^2 B_k \left(1+\frac{B_k-1}{B_\setS(\vecw_{N_{k-1}} )} \right) M^2(\vecw_{N_{k-1}}) \\
\ge & (1 - 2\gamma \lambda B_k) \twonms{ \vecw_{N_{k-1}} - \vecw^*}^2 + \frac{1}{2}\gamma^2 M^2 B_k \left(1+\frac{B_k-1}{B_\setS(\vecw_{N_{k-1}})} \right).
\end{align*}
Then, we take expectation over the randomness of the whole algorithm and obtain
\begin{align*}
\EXPS{\twonms{\vecw_{N_k} - \vecw^* }^2} \ge & (1 - 2\gamma \lambda B_k) \EXPS{\twonms{ \vecw_{N_{k-1}} - \vecw^*}^2} + \frac{1}{2}\gamma^2 M^2 B_k \left(1+(B_k-1)\EXP{\frac{1}{B_\setS(\vecw_{N_{k-1}})}} \right) \\
\ge & (1 - 2\gamma \lambda B_k) \EXPS{\twonms{ \vecw_{N_{k-1}} - \vecw^*}^2} + \frac{1}{2}\gamma^2 M^2 B_k \left(1+(B_k-1)\frac{1}{\EXPS{B_\setS(\vecw_{N_{k-1}})} }\right)  \\
\ge & (1 - 2\gamma \lambda B_k) \EXPS{\twonms{ \vecw_{N_{k-1}} - \vecw^*}^2} + \frac{1}{2}(1+\delta)\gamma^2 M^2 B_k,
\end{align*}
where the second inequality is due to Jensen's inequality, and the third inequality is due to the fact that $B_k \ge 1+\delta \EXPS{B_\setS(\vecw_{N_{k-1}})}$.

Rolling out the above recursion, and denoting $\alpha_k = 2\gamma \lambda B_k\in[0,1]$, we have
\begin{align*}
\EE \left[\| \vecw_{N_K} - \vecw^* \|^2_2 \right] &\geq \| \vecw_0 - \vecw^* \|_2^2 \left( \prod_{k=1}^K (1 - \alpha_k) \right) + \frac{1}{2}(1 + \delta) \gamma^2 M^2  \left[B_K + \sum_{k=1}^{K-1} \prod_{i=k+1}^{K} (1 - \alpha_i) B_k \right] \\
&= \| \vecw_0 - \vecw^* \|_2^2 \left( \prod_{i=1}^K (1 - \alpha_i) \right) + \frac{1}{4} (1 + \delta) \frac{\gamma M^2}{\lambda}  \left[\alpha_K + \sum_{k=1}^{K-1} \prod_{i=k+1}^{K} (1 - \alpha_i) \alpha_k \right].
\end{align*} 
Now the number of gradient updates is given by $\sum_{k=1}^K B_k = T$, and consequently, $\sum_{k=1}^K \alpha_k = 2\gamma \lambda T$. Since we consider the case when $T \geq \frac{c}{\gamma \lambda}$ for some universal constant $c>0$ (and SGD only converges in this regime), so we have $\sum_{k=1}^K \alpha_k \geq 2c$.

Substituting the value of step-size $\gamma$, we see that in order to complete the proof, it suffices to show that the quantity
\begin{align*}
J(\alpha) = \alpha_K + \sum_{k=1}^{K-1} \prod_{i=k+1}^{K} (1 - \alpha_i) \alpha_k
\end{align*}
is lower bounded as $\Omega(1)$. In order to show this, note that $J(\alpha)$ can be equivalently expressed as the CDF of a geometric distribution with non-uniform probabilities of success $\alpha_k$. We could further see that
$$
J(\alpha) = 1 - \prod_{k=1}^K(1-\alpha_k) \ge 1-\left[ \frac{1}{K}\sum_{k=1}^K(1-\alpha_k)\right]^K \ge 1-(1-2c/K)^K,
$$
and the last term is lower bounded by a constant for all $K \geq 1$.

\subsection{Necessity of $B\le \bigo(\frac{1}{\lambda \gamma})$}\label{prf:strong_smooth_lower}
In this section, we show that, up to a constant factor, the condition $B\le \frac{1}{2\gamma \lambda}$ in Theorem~\ref{thm:convergence_strong}~and~\ref{thm:lowerboundvarB}, is actually necessary for mini-batch SGD to converge when $F(\vecw)$ is strongly convex. More precisely, we can show that, when $B > \frac{2}{\gamma \lambda}$, mini-batch SGD diverges.

\begin{theorem}\label{thm:strong_smooth_lower}
Suppose that $F(\vecw)$ is $\lambda$-strongly convex. Condition on the model parameter $\vecw_{kB}$ obtained after $k$ iterations. Suppose that $\vecw_{kB}-\gamma\sum_{i\in\mathcal{I}}\gradf_i(\vecw_{kB})\in\W$ for all $\mathcal{I}\in[n]^B$. Then, if $B > \frac{2}{\gamma \lambda}$, we have
$$
\EE \left[ \| \vecw_{(k+1)B} - \vecw^* \|_2\mid \vecw_{kB} \right] > \| \vecw_{kB} - \vecw^* \|_2.
$$
\end{theorem}
\begin{proof}
We have
\begin{align*}
\EE \left[ \| \vecw_{(k+1)B} - \vecw_{kB} \|_2 \mid \vecw_{kB} \right] &\geq \twonm{ \EXPS{\vecw_{(k+1)B} - \vecw_{kB} \mid \vecw_{kB}} } \\
&= \gamma \twonm{ \sum_{t=kB}^{(k+1)B-1} \EXPS{ \nabla f_{s_t}(\vecw_{kB}) \mid \vecw_{kB}} } \\
&= \gamma B \| \nabla F(\vecw_{kB})\|_2 \\
&\geq \gamma B \lambda \| \vecw_{kB} - \vecw^* \|_2,
\end{align*}
where the first step follows by Jensen's inequality, and the last by strong convexity.

This allows us to conclude that if $B > \frac{2}{\gamma \lambda}$, $
\EE \left[ \| \vecw_{(k+1)B} - \vecw_{kB} \|_2 \mid \vecw_{kB} \right] > 2 \| \vecw_{kB} - \vecw^* \|_2.
$ 
Then, by triangle inequality, 
$$
\EE \left[ \| \vecw_{(k+1)B} - \vecw^* \|_2 \mid \vecw_{kB} \right] \ge \EE \left[ \| \vecw_{(k+1)B} - \vecw_{kB} \|_2 \mid \vecw_{kB} \right] - \| \vecw_{kB} - \vecw^* \|_2 > \| \vecw_{kB} - \vecw^* \|_2,
$$
and thus mini-batch SGD diverges.
\end{proof}

%% file: proof_diversity_inducing.tex
\section{Proof of Theorem~\ref{thm:DIM}}\label{prf:DIM}
For dropout, we have
\begin{equation}\label{eq:dropout}
\begin{aligned}
B_\setS^{\sf{drop}}(\vecw) &= n \frac{ \sum_{i=1}^n\EXPS{\twonms{\mat{D}_i\gradf_i(\vecw)}^2} }{ \EXPS{ \twonms{\sum_{i=1}^n\mat{D}_i\gradf_i(\vecw)}^2 } } \\
&= \frac{ n \sum_{i=1}^n (1-p)\twonms{\gradf_i(\vecw)}^2 }{  \sum_{i=1}^n(1-p)\twonms{\gradf_i(\vecw)}^2 + (1-p)^2\sum_{j\neq k}\innerps{\gradf_j(\vecw)}{\gradf_k(\vecw)}  }.
\end{aligned}
\end{equation}
Recall that 
$$
B_\setS(\vecw) = \frac{ n \sum_{i=1}^n \twonms{\gradf_i(\vecw)}^2 }{  \sum_{i=1}^n\twonms{\gradf_i(\vecw)}^2 + \sum_{j\neq k}\innerps{\gradf_j(\vecw)}{\gradf_k(\vecw)}  },
$$
and we can see that for any $\vecw$ such that $\sum_{j\neq k}\innerps{\gradf_j(\vecw)}{\gradf_k(\vecw)}   \ge 0$, we must have
$B_\setS(\vecw) \le n$. In this case, we have
$$
B_\setS^{\sf{drop}}(\vecw)\ge \frac{n \sum_{i=1}^n (1-p)\twonms{\gradf_i(\vecw)}^2 }{  \sum_{i=1}^n(1-p)\twonms{\gradf_i(\vecw)}^2 + (1-p)\sum_{j\neq k}\innerps{\gradf_j(\vecw)}{\gradf_k(\vecw)} }=B_\setS(\vecw).
$$
On the other hand, if $\sum_{j\neq k}\innerps{\gradf_j(\vecw)}{\gradf_k(\vecw)}  < 0$, we must have $B_\setS(\vecw) > n$, and one can simply check that we also have $B_\setS^{\sf{drop}}(\vecw) > n$.

For stochastic gradient Langevin dynamics, we have
\begin{equation}\label{eq:sgld}
B_\setS^{\sf{sgld}}(\vecw) =\frac{ n\sum_{i=1}^n\EXPS{\twonms{\gradf_i(\vecw)+\vect{\xi}_i}^2} }{ \EXPS{ \twonms{\sum_{i=1}^n(\gradf_i(\vecw)+\vect{\xi}_i)}^2 } } =\frac{ n \sum_{i=1}^n \twonms{\gradf_i(\vecw)}^2+n^2d\sigma^2 }{  \twonms{\sum_{i=1}^n \gradf_i(\vecw)}^2+ nd\sigma^2  }.
\end{equation}

Therefore, as long as $B_\setS(\vecw) = \frac{ n\sum_{i=1}^n \twonms{\gradf_i(\vecw)}^2 }{\twonms{\sum_{i=1}^n \gradf_i(\vecw)}^2} \le n$, we have $B_\setS^{\sf{sgld}}(\vecw) \ge B_\setS(\vecw)$. In addition, if $B_\setS(\vecw) > n$, then $B_\setS^{\sf{sgld}}(\vecw) > n$.

For quantization, one can simply check that for any $i\in[n]$, we have $\EXPS{\twonms{Q(\gradf_i(\vecw))}^2} = \twonms{\gradf_i(\vecw)}\| \gradf_i(\vecw) \|_1$, and for any $j\neq k$, we have $\EXPS{\innerps{Q(\gradf_j(\vecw))}{Q(\gradf_k(\vecw))}} = \innerps{\gradf_j(\vecw)}{\gradf_k(\vecw)}$. Consequently,
\begin{equation}\label{eq:quant}
\begin{aligned}
B_\setS^{\sf{quant}}(\vecw) &= \frac{ n\sum_{i=1}^n\EXPS{\twonms{Q(\gradf_i(\vecw))}^2} }{ \EXPS{ \twonms{\sum_{i=1}^nQ(\gradf_i(\vecw))}^2 } } \\
& =   \frac{ n\sum_{i=1}^n\twonms{\gradf_i(\vecw)}\| \gradf_i(\vecw) \|_1  }{ \sum_{i=1}^n\twonms{\gradf_i(\vecw)}\| \gradf_i(\vecw) \|_1 + \sum_{j\neq k}\innerps{\gradf_j(\vecw)}{\gradf_k(\vecw)}  }.
\end{aligned}
\end{equation}
We define
$$
\Delta_\setS^{\sf{quant}} (\vecw) := \frac{ \sum_{i=1}^n\twonms{\gradf_i(\vecw)}\| \gradf_i(\vecw) \|_1  }{ \sum_{i=1}^n\twonms{\gradf_i(\vecw)}\| \gradf_i(\vecw) \|_1 + \sum_{j\neq k}\innerps{\gradf_j(\vecw)}{\gradf_k(\vecw)}  },
$$
and
$$
\Delta_\setS(\vecw) := \frac{ \sum_{i=1}^n\twonms{\gradf_i(\vecw)}^2  }{ \sum_{i=1}^n \twonms{\gradf_i(\vecw)}^2 + \sum_{j\neq k}\innerps{\gradf_j(\vecw)}{\gradf_k(\vecw)}  },
$$
and we have $B_\setS^{\sf{quant}}(\vecw) = n\Delta_\setS^{\sf{quant}} (\vecw)$ and $B_\setS = n\Delta_\setS (\vecw)$. One can now check that due to the fact that $\|\vect{v}\|_2 \|\vect{v} \|_1 \geq \|\vect{v}\|_2^2$ for any vector $\vect{v}$, when $\Delta_\setS (\vecw)\in(0,1)$, we have $\Delta_\setS^{\sf{quant}} (\vecw) > \Delta_\setS (\vecw)$, and when $\Delta_\setS (\vecw) > 1$, we have $\Delta_\setS^{\sf{quant}} (\vecw) > 1$.

%% file: proof_stability.tex
\section{Stability}
\subsection{Notation}
To assist the demonstration of the proof of our stability results, we define the following quantities. Let
$$
\overline{M}^2(\vecw,\vecw'):=\frac{1}{n}\sum_{i=1}^n\twonms{\gradf_i(\vecw) - \gradf_i(\vecw')}^2 \quad\text{and}\quad \overline{G}(\vecw,\vecw'):=\twonms{\gradF(\vecw) - \gradF(\vecw')}^2.
$$
One can see that $\overline{B}_\setS(\vecw,\vecw') = \frac{\overline{M}^2(\vecw,\vecw')}{ \overline{G}(\vecw,\vecw')}$. We also define
$$
\overline{B}_\setS = \inf_{\vecw\neq\vecw'} \overline{B}_\setS(\vecw,\vecw').
$$

\subsection{Proof of Theorem~\ref{thm:stab_cvx} and Corollary~\ref{cor:gen_cvx}}\label{prf:stab_cvx}
We first recall the problem setting. Suppose that there are two sample sets $\setS$ and $\setS^{(I)}$ which differs at one data point located at a random position $I$, which is uniformly distributed in $[n]$. We run the same (projected) parallel mini-batch SGD on both data sets, and after the $k$-th parallel iteration, we obtain $\vecwb_{kB}$ and $\tvecwb_{kB}$, respectively. After a total number of $T$ gradient updates, \ie $T/B$ parallel iterations, we obtain $\vecw_T$ and $\tvecw_T$. Let $s_t$, $t=0,1,\ldots, T-1$ be the sequence of indices of samples used by the algorithm. In our setting, $s_t$ are i.i.d. uniformly distributed in $\{1,2,\ldots, n\}$. Let $\vecz_{s_t}\in\setS$ and $\widetilde{\vecz}_{s_t}\in\setS^{(I)}$, $t=0,\ldots, T-1$ be the data point used in the algorithms running on the two data sets, respectively. Then, we know that with probability $1-\frac{1}{n}$, $\vecz_{s_t} = \widetilde{\vecz}_{s_t}$, and with probability $\frac{1}{n}$, $\vecz_{s_t} \neq\widetilde{\vecz}_{s_t}$. We simplify the notations of the risk function associated with $\vecz_{s_t}$ and $\widetilde{\vecz}_{s_t}$ by $f_{s_t}(\vecw):=f(\vecw;\vecz_{s_t})$, and $\tf_{s_t}(\vecw):=f(\vecw;\widetilde\vecz_{s_t})$, respectively.

We now prove Theorem~\ref{thm:stab_cvx}. Throughout this proof, we only consider the case where $B>1$ and omit the indicator function $\indi_{B>1}$.
We condition on the data sets and the event that the choice of $\gamma$ is ``good'', as shown in~\eqref{eq:stab_stepsize}. Specifically, we condition on the samples $\setS$ and $\setS^\prime$, and the event $\Gamma$:
\begin{equation}\label{eq:good_gamma}
\Gamma =  \left\{ \gamma \le \frac{2}{\beta(1+\frac{1}{n-1} + \frac{B-1}{\overline{B}_\setS})} \right\} = \left\{ \overline{B}_\setS \ge \frac{B-1}{\frac{2}{\gamma \beta} - 1 - \frac{1}{n-1}} \right\}.
\end{equation}

Our goal is to bound $\abs{ \epsilon_{\text{stab}}(\setS, \setS')}$. 
Since we assume that $f(\vecw;\vecz)$ is $L$-Lipschitz on $\W$, we have
\begin{equation}\label{eq:stab-model-dist}
\abs{\epsilon_{\text{stab}}(\setS, \setS')} \le L \E_{I,A \mid \Gamma} \left[ \twonms{A(\setS^{(I)})-A(\setS)} \right] = L \E_{I,A \mid \Gamma} \left[ \twonms{\vecwb_{T} - \tvecwb_T} \right],
\end{equation}
and thus it suffices to bound $\E_{I,A \mid \Gamma} \left[ \twonms{\vecwb_{T} - \tvecwb_T} \right]$.

Consider the samples used in the $(k+1)$-th parallel iteration in the two algorithm instances, \ie $\{\vecz_{s_t}\}_{t=kB}^{(k+1)B-1}$, and $\{\widetilde{\vecz}_{s_t}\}_{t=kB}^{(k+1)B-1}$. Let $H_{k+1}$ be the number of instances that the two data points with the same index being different in these two sample sets (i.e., $s_t=I$). According to our sampling scheme, $H_{k+1}\sim\text{bin}(B,\frac{1}{n})$. We condition on the event that $H_{k+1} = h$. Without loss of generality, we assume that $\vecz_{s_t} = \tvecz_{s_t}$ for all $t=kB,\ldots, (k+1)B-h-1$, and $\vecz_{s_t} \neq \tvecz_{s_t}$ for all $t=(k+1)B-h,\ldots,(k+1)B-1$. Consider the first $B-h$ terms. For the unconstrained optimization, we have
\begin{equation}\label{eq:firstb_h}
\twonms{\vecwb_{(k+1)B-h} - \tvecwb_{(k+1)B-h}}^2 = \twonms{ ( \vecwb_{kB} - \gamma\sum_{t=kB}^{(k+1)B-h-1} \gradf_{s_t}(\vecwb_{kB})) - ( \tvecwb_{kB} - \gamma\sum_{t=kB}^{(k+1)B-h-1} \nabla \tf_{s_t}(\tvecwb_{kB})) }^2.
\end{equation}
For the algorithm with projection, the $B$ gradient update steps are the same as the unconstrained algorithm, and projection step is conducted once all the gradient updates are finished. Therefore,~\eqref{eq:firstb_h} also holds for projected algorithm.

Since $f_{s_t}(\vecw) = \tf_{s_t}(\vecw)$ for all $t=kB,\ldots, (k+1)B-h-1$, we further have
\begin{equation}\label{eq:stab_split}
\begin{aligned}
& \twonms{\vecwb_{(k+1)B-h} - \tvecwb_{(k+1)B-h}}^2 \\
 =& \twonms{\vecwb_{kB}-\tvecwb_{kB}}^2 - 2 \innerps{\vecwb_{kB} - \tvecwb_{kB}}{\gamma \sum_{t=kB}^{(k+1)B-h-1} \gradf_{s_t}(\vecwb_{kB}) - \gradf_{s_t}(\tvecwb_{kB}) } \\
&+\gamma^2 \twonms{\sum_{t=kB}^{(k+1)B-h-1} \gradf_{s_t}(\vecwb_{kB}) - \gradf_{s_t}(\tvecwb_{kB})}^2 \\
= &  \twonms{\vecwb_{kB}-\tvecwb_{kB}}^2 - 2  \innerps{\vecwb_{kB} - \tvecwb_{kB}}{\gamma \sum_{t=kB}^{(k+1)B-h-1} \gradf_{s_t}(\vecwb_{kB}) - \gradf_{s_t}(\tvecwb_{kB}) } \\
&+ \gamma^2 \sum_{t= kB}^{(k+1)B-h-1} \twonms{ \gradf_{s_t}(\vecwb_{kB}) - \gradf_{s_t}(\tvecwb_{kB}) }^2  \\
&+ 2\gamma^2 \sum_{i = kB}^{(k+1)B-h-1} \sum_{j=i+1}^{(k+1)B-h-1} \innerps{ \gradf_{s_i}(\vecwb_{kB}) - \gradf_{s_i}(\tvecwb_{kB}) }{  \gradf_{s_j}(\vecwb_{kB}) - \gradf_{s_j}(\tvecwb_{kB})  }.
\end{aligned}
\end{equation}
We denote the sequence of indices selected by the mini-batch SGD algorithm up to the $t$-th sampled data point as $A_t$, i.e., $A_t=\{s_0,\ldots,s_{t-1}\}$. In the following steps, we condition on $A_{kB}$ and the event that $H_{k+1}=h$, and take expectation over the randomness of the SGD algorithm in the $(k+1)$-th parallel iteration and the random choice of $I$. 

We consider each term in~\eqref{eq:stab_split}. For the term $\twonms{ \gradf_{s_t}(\vecwb_{kB}) - \gradf_{s_t}(\tvecwb_{kB}) }^2$, conditioned on the event that $\vecz_{s_t} = \tvecz_{s_t}$, we know that $s_t$ is uniformly distributed in $[n]\setminus\{I\}$. Since $I$ is uniformly distributed in $[n]$, we know that the marginal distribution of $s_t$ is uniform in $[n]$. We have
$$
\E_{I,A\mid H_{k+1}, A_{kB}, \Gamma}[ \twonms{ \gradf_{s_t}(\vecwb_{kB}) - \gradf_{s_t}(\tvecwb_{kB}) }^2 ] = \overline{M}^2(\vecwb_{kB}, \tvecwb_{kB}).
$$

Then we find the conditional expectation of $\innerps{ \gradf_{s_i}(\vecwb_{kB}) - \gradf_{s_i}(\tvecwb_{kB}) }{  \gradf_{s_j}(\vecwb_{kB}) - \gradf_{s_j}(\tvecwb_{kB})  }$. We have the following lemma.
\begin{lemma}\label{lem:distribution}
For any $i,j$ such that $kB \le i,j \le (k+1)B-h-1$ and $i\neq j$, we have
\begin{equation}\label{eq:exp_si_sj}
\begin{aligned}
& \E_{I,A\mid H_{k+1}, A_{kB}, \Gamma}[\innerps{ \gradf_{s_i}(\vecwb_{kB}) - \gradf_{s_i}(\tvecwb_{kB}) }{  \gradf_{s_j}(\vecwb_{kB}) - \gradf_{s_j}(\tvecwb_{kB})  }] \\
= &\frac{1}{(n-1)^2}\overline{M}^2(\vecwb_{kB}, \tvecwb_{kB}) + \frac{n(n-2)}{(n-1)^2} \overline{G}(\vecwb_{kB}, \tvecwb_{kB}).
\end{aligned}
\end{equation}
\end{lemma}
We prove Lemma~\ref{lem:distribution} in Appendix~\ref{prf:distribution}. According to this lemma, we have
\begin{equation}\label{eq:exp_stab_split}
\begin{aligned}
& \E_{I,A\mid H_{k+1}, A_{kB}, \Gamma}[\twonms{\vecwb_{(k+1)B-h} - \tvecwb_{(k+1)B-h}}^2] \\
= & \twonms{\vecwb_{kB}-\tvecwb_{kB}}^2 - 2\E_{I,A\mid H_{k+1}, A_{kB}, \Gamma} [\innerps{\vecwb_{kB} - \tvecwb_{kB}}{\gamma \sum_{t=kB}^{(k+1)B-h-1} \gradf_{s_t}(\vecwb_{kB}) - \gradf_{s_t}(\tvecwb_{kB}) }] \\
&+ \gamma^2 (B-h) \overline{M}^2(\vecwb_{kB}, \tvecwb_{kB}) \\
&+ \gamma^2 (B-h)(B-h-1) \left[ \frac{1}{(n-1)^2}\overline{M}^2(\vecwb_{kB}, \tvecwb_{kB}) + \frac{n(n-2)}{(n-1)^2} \overline{G}(\vecwb_{kB}, \tvecwb_{kB}) \right]  \\
\le & \twonms{\vecwb_{kB}-\tvecwb_{kB}}^2 - 2 \E_{I,A\mid H_{k+1}, A_{kB}, \Gamma} [\innerps{\vecwb_{kB} - \tvecwb_{kB}}{\gamma \sum_{t=kB}^{(k+1)B-h-1} \gradf_{s_t}(\vecwb_{kB}) - \gradf_{s_t}(\tvecwb_{kB}) }] \\
&+ \gamma^2 (B-h) \overline{M}^2(\vecwb_{kB}, \tvecwb_{kB}) + \gamma^2 (B-h) \left[ \frac{1}{n-1}\overline{M}^2(\vecwb_{kB}, \tvecwb_{kB}) + (B-1)\frac{\overline{M}^2(\vecwb_{kB}, \tvecwb_{kB})}{\overline{B}_\setS(\vecwb_{kB}, \tvecwb_{kB})} \right]  \\
\le & \twonms{\vecwb_{kB}-\tvecwb_{kB}}^2 - 2 \gamma \sum_{t=kB}^{(k+1)B-h-1} \E_{I,A\mid H_{k+1}, A_{kB}, \Gamma} [\innerps{\vecwb_{kB} - \tvecwb_{kB}}{ \gradf_{s_t}(\vecwb_{kB}) - \gradf_{s_t}(\tvecwb_{kB}) }] \\
&+ \gamma^2 (B-h) ( 1+ \frac{1}{n-1}+ \frac{B-1}{\overline{B}_\setS} )\overline{M}^2(\vecwb_{kB}, \tvecwb_{kB}).
\end{aligned}
\end{equation}

By the co-coercive property of convex and smooth functions, we know that
$$
\innerps{\vecwb_{kB} - \tvecwb_{kB}}{  \gradf_{s_t}(\vecwb_{kB}) - \gradf_{s_t}(\tvecwb_{kB}) } \ge \frac{1}{\beta} \twonms{\gradf_{s_t}(\vecwb_{kB}) - \gradf_{s_t}(\tvecwb_{kB})}^2.
$$

Then, we obtain
\begin{equation}\label{eq:exp_split2}
\begin{aligned}
\E_{I,A\mid H_{k+1}, A_{kB}, \Gamma}&[\twonms{\vecwb_{(k+1)B-h} - \tvecwb_{(k+1)B-h}}^2] \le  \twonms{\vecwb_{kB}-\tvecwb_{kB}}^2\\
& - (2\frac{\gamma}{\beta}-\gamma^2(1+\frac{1}{n-1}+\frac{B-1}{\overline{B}_\setS}))(B-h)\overline{M}^2(\vecwb_{kB}, \tvecwb_{kB}).
\end{aligned}
\end{equation}
Since we condition on good choice of $\gamma$ in~\eqref{eq:good_gamma}, we have
$$
\E_{I,A\mid H_{k+1}, A_{kB}, \Gamma}[\twonms{\vecwb_{(k+1)B-h} - \tvecwb_{(k+1)B-h}}^2] \le \twonms{\vecwb_{kB}-\tvecwb_{kB}}^2.
$$
Then by Jensen's inequality, we have
\begin{equation}\label{eq:first_b_h}
\E_{I,A\mid H_{k+1}, A_{kB}, \Gamma}[\twonms{\vecwb_{(k+1)B-h} - \tvecwb_{(k+1)B-h}}] \le \twonms{\vecwb_{kB}-\tvecwb_{kB}}.
\end{equation}
For the last $h$ terms, since the loss functions are all $L$-Lipschitz, we obtain
\begin{equation}\label{eq:exp_scvx_last_h}
\twonms{\vecwb_{(k+1)B} - \tvecwb_{(k+1)B}}  \le \twonms{\vecwb_{(k+1)B-h} - \tvecwb_{(k+1)B-h}} + 2\gamma L h.
\end{equation}
Then, combining with equation~\eqref{eq:first_b_h}, we have
\begin{equation}\label{eq:exp_scvx_b}
\E_{I,A\mid H_{k+1}, A_{kB}, \Gamma}[\twonms{\vecwb_{(k+1)B} - \tvecwb_{(k+1)B}}] \le \twonms{\vecwb_{kB}-\tvecwb_{kB}} + 2\gamma L h.
\end{equation}
Taking expectation over $H_{k+1}$ yields
$$
\E_{I,A\mid A_{kB}, \Gamma}[\twonms{\vecwb_{(k+1)B} - \tvecwb_{(k+1)B}}] \le \twonms{\vecwb_{kB}-\tvecwb_{kB}} + 2\gamma L \frac{B}{n}.
$$
Then we take expectation over the randomness of the first $k$ parallel iterations and obtain
\begin{equation}\label{eq:exp_i_a}
\E_{I,A \mid \Gamma}[\twonms{\vecwb_{(k+1)B} - \tvecwb_{(k+1)B}}] \le \E_{I,A\mid \Gamma} [\twonms{\vecwb_{kB}-\tvecwb_{kB}}] + 2\gamma L \frac{B}{n}.
\end{equation}
Summing up~\eqref{eq:exp_i_a} for $k= 0, 1,\ldots, \frac{T}{B}-1$ and taking expectation over the data sets, we have
\begin{equation}\label{eq:exp_i_a2}
\E_{I,A\mid \Gamma}[\twonms{\vecwb_{T} - \tvecwb_{T}}] \le 2\gamma L \frac{T}{n}.
\end{equation}
Combining equations~\eqref{eq:stab-model-dist} and~\eqref{eq:exp_i_a2}, we complete the proof of Theorem~\ref{thm:stab_cvx}, \ie
when $\Gamma$ happens,
\begin{equation}\label{eq:eps_stab_gamma}
\abs{\epsilon_{\text{stab}}(\setS, \setS')} \le L \E_{I,A \mid \Gamma} \left[ \twonms{\vecwb_{T} - \tvecwb_T} \right] \le 2\gamma L^2 \frac{T}{n}.
\end{equation}
To prove Corollary~\ref{cor:gen_cvx}, we notice the fact that when $\Gamma$ does not happen, we simply have
\begin{equation}\label{eq:exp_i_a3}
\abs{\epsilon_{\text{stab}}(\setS, \setS')} \le L\E_{I,A\mid \bar{\Gamma}}[\twonms{\vecwb_{T} - \tvecwb_{T}}] \le 2\gamma L^2 T.
\end{equation}
According to~\eqref{eq:def_eta}, $\eta=\probs{\bar{\Gamma}}$. Then, according to~\eqref{eq:eps_stab_gamma} and~\eqref{eq:exp_i_a3}, we get
\begin{align*}
\epsilon_{\text{gen}} &\le \E_{\setS, \setS' \mid \Gamma} \left[ \abs{\epsilon_{\text{stab}}(\setS, \setS')}  \right] \probs{\Gamma} +  \E_{\setS, \setS' \mid \bar{\Gamma}} \left[ \abs{\epsilon_{\text{stab}}(\setS, \setS')}  \right] \probs{\bar{\Gamma}} \\
 &\le  2\gamma L^2 \frac{T}{n}(1-\eta) + 2\gamma L^2 T\eta,
\end{align*}
which completes the proof.

\subsection{Proof of Theorem~\ref{thm:stab_strong_cvx} and Corollary~\ref{cor:gen_strong_cvx}}\label{prf:stab_strong_cvx}
The proof of Theorem~\ref{thm:stab_strong_cvx} follows an argument similar to the proof of Theorem~\ref{thm:stab_cvx}. We define the event $\Gamma$ that the step size is ``good'' in the following way, as shown in~\eqref{eq:stab_strong_stepsize} (slightly different from the convex risk functions):
\begin{equation}\label{eq:good_gamma_strong}
\Gamma = \left\{ \gamma \le \frac{2}{(\beta+\lambda)(1+\frac{1}{n-1} + \frac{B-1}{\overline{B}_\setS})} \right\} = \left\{ \overline{B}_\setS \ge \frac{B-1}{\frac{2}{\gamma (\beta+\lambda)} - 1 - \frac{1}{n-1}} \right\}.
\end{equation}
To prove Theorem~\ref{thm:stab_strong_cvx}, our goal is still to bound $\E_{I,A \mid \Gamma} \left[ \twonms{\vecwb_{T} - \tvecwb_T} \right]$.
Since the result in~\eqref{eq:exp_stab_split} still holds for strongly convex functions, we can obtain
\begin{equation}\label{eq:exp_stab_strong_split}
\begin{aligned}
& \E_{I,A\mid H_{k+1}, A_{kB}, \Gamma}[\twonms{\vecwb_{(k+1)B-h} - \tvecwb_{(k+1)B-h}}^2] \\
\le & \twonms{\vecwb_{kB}-\tvecwb_{kB}}^2 - 2 \gamma \sum_{t=kB}^{(k+1)B-h-1} \E_{I,A\mid H_{k+1}, A_{kB}, \Gamma} [ \innerps{\vecwb_{kB} - \tvecwb_{kB}}{  \gradf_{s_t}(\vecwb_{kB}) - \gradf_{s_t}(\tvecwb_{kB}) }] \\
&+ \gamma^2 (B-h) ( 1+ \frac{1}{n-1}+ \frac{B-1}{\overline{B}_\setS} )\overline{M}^2(\vecwb_{kB}, \tvecwb_{kB}),
\end{aligned}
\end{equation}
where $H_{k+1}$ is defined in the same way as in the proof of Theorem~\ref{thm:stab_cvx}. For strongly convex functions, we have the following co-coercive property:
$$
 \innerps{\vecwb_{kB} - \tvecwb_{kB}}{  \gradf_{s_t}(\vecwb_{kB}) - \gradf_{s_t}(\tvecwb_{kB}) } \ge \frac{\beta \lambda}{\beta + \lambda} \twonms{\vecwb_{kB} - \tvecwb_{kB}}^2 + \frac{1}{\beta+\lambda}\twonms{\gradf_{s_t}(\vecwb_{kB}) - \gradf_{s_t}(\tvecwb_{kB})}^2,
$$
which gives us
\begin{equation}\label{eq:exp_stab_strong_split2}
\begin{aligned}
\E_{I,A\mid H_{k+1}, A_{kB}, \Gamma}&[\twonms{\vecwb_{(k+1)B-h} - \tvecwb_{(k+1)B-h}}^2] \le \left( 1-2\gamma(B-h)\frac{\beta \lambda}{\beta + \lambda} \right)\twonms{\vecwb_{kB}-\tvecwb_{kB}}^2 \\
&-\gamma (B-h)\left[ \frac{2}{\beta + \lambda}-\gamma(1+\frac{1}{n-1}+\frac{B-1}{\overline{B}_\setS}) \right]\overline{M}^2(\vecwb_{kB}, \tvecwb_{kB}).
\end{aligned}
\end{equation}
Since we only consider the regime where $B \le \frac{1}{2\gamma \lambda}$, one can check that $1-2\gamma(B-h)\frac{\beta \lambda}{\beta + \lambda}>0$ for any $h=0,\ldots, B$. Conditioned on the data sets and the good choice of $\gamma$, we have
\begin{equation}\label{eq:strong_first_b_h}
\E_{I,A\mid H_{k+1}, A_{kB}, \Gamma}[\twonms{\vecwb_{(k+1)B-h} - \tvecwb_{(k+1)B-h}}^2] \le \left( 1-2\gamma B\frac{\beta \lambda}{\beta + \lambda} \right)\twonms{\vecwb_{kB}-\tvecwb_{kB}}^2.
\end{equation}
With Jensen's inequality and the fact that $\sqrt{1-x} \le 1-\frac{x}{2}$ for any $x\in[0,1]$, we have
\begin{equation}\label{eq:strong_first_b_h_2}
\E_{I,A\mid H_{k+1}, A_{kB}, \Gamma}[\twonms{\vecwb_{(k+1)B-h} - \tvecwb_{(k+1)B-h}}] \le \left( 1-\gamma B\frac{\beta \lambda}{\beta + \lambda} \right)\twonms{\vecwb_{kB}-\tvecwb_{kB}}.
\end{equation}
For the last $h$ terms, we have
\begin{equation}\label{eq:strong_last_h}
\begin{aligned}
\E_{I,A\mid H_{k+1}, A_{kB}, \Gamma} [\twonms{\vecwb_{(k+1)B} - \tvecwb_{(k+1)B}}] \le \twonms{\vecwb_{(k+1)B-h} - \tvecwb_{(k+1)B-h}} + 2\gamma L h.
\end{aligned}
\end{equation}
Combined with equation~\eqref{eq:strong_first_b_h_2}, we obtain
$$
\E_{I,A\mid H_{k+1}, A_{kB}, \Gamma}[\twonms{\vecwb_{(k+1)B} - \tvecwb_{(k+1)B}}] \le \left( 1-\gamma B\frac{\beta \lambda}{\beta + \lambda} \right)\twonms{\vecwb_{kB}-\tvecwb_{kB}} + 2\gamma L h,
$$
and by taking expectation over $h$ we have
$$
\E_{I,A\mid A_{kB}, \Gamma}[\twonms{\vecwb_{(k+1)B} - \tvecwb_{(k+1)B}}] \le \left( 1-\gamma B\frac{\beta \lambda}{\beta + \lambda} \right)\twonms{\vecwb_{kB}-\tvecwb_{kB}} + 2\gamma L \frac{B}{n}.
$$
Taking expectation over $A_{kB}$ yields
\begin{equation}\label{eq:iter-gen-strong}
\E_{I,A\mid  \Gamma}[\twonms{\vecwb_{(k+1)B} - \tvecwb_{(k+1)B}}] \le \left( 1-\gamma B\frac{\beta \lambda}{\beta + \lambda} \right)\E_{I,A\mid  \Gamma}[\twonms{\vecwb_{kB}-\tvecwb_{kB}}] + 2\gamma L \frac{B}{n}.
\end{equation}
Iterating equation~\eqref{eq:iter-gen-strong} yields
\begin{equation}\label{eq:exp_strong_1}
\E_{I,A\mid \Gamma}[\twonms{\vecwb_{T} - \tvecwb_{T}}] \le \frac{4L}{\lambda n}.
\end{equation}
Combining equations~\eqref{eq:stab-model-dist} and~\eqref{eq:exp_strong_1}, we prove Theorem~\ref{thm:stab_strong_cvx}, \ie when $\Gamma$ happens,
\begin{equation}\label{eq:eps_stab_gamma_strong}
\abs{\epsilon_{\text{stab}}(\setS, \setS')} \le L \E_{I,A \mid \Gamma} \left[ \twonms{\vecwb_{T} - \tvecwb_T} \right] \le \frac{4L^2}{\lambda n}.
\end{equation}
To prove Corollary~\ref{cor:gen_strong_cvx}, we notice the fact that,
when $\Gamma$ does not occur, we simply have
\begin{equation}\label{eq:exp_strong_2}
\abs{\epsilon_{\text{stab}}(\setS, \setS')} \le L\E_{I,A\mid \bar{\Gamma}}[\twonms{\vecwb_{T} - \tvecwb_{T}}] \le 2\gamma L^2 T.
\end{equation}
According to~\eqref{eq:def_eta_strong}, $\eta=\probs{\bar{\Gamma}}$. Then, according to~\eqref{eq:eps_stab_gamma_strong} and~\eqref{eq:exp_strong_2}, we get
\begin{align*}
\epsilon_{\text{gen}} &\le \E_{\setS, \setS' \mid \Gamma} \left[ \abs{\epsilon_{\text{stab}}(\setS, \setS')}  \right] \probs{\Gamma} +  \E_{\setS, \setS' \mid \bar{\Gamma}} \left[ \abs{\epsilon_{\text{stab}}(\setS, \setS')}  \right] \probs{\bar{\Gamma}} \\
 &\le  \frac{4L^2}{\lambda n}(1-\eta) + 2\gamma L^2 T\eta,
\end{align*}
which completes the proof.

\subsection{Proof of Lemma~\ref{lem:distribution}}\label{prf:distribution}
One can interpret $\overline{M}^2(\vecw,\vecw^\prime)$ and $\overline{G}(\vecw,\vecw^\prime)$ as follows. Let $\mathcal{P}_1$ be a distribution on $[n]\times[n]$ with PMF \begin{equation}\label{eq:dist_p1}
p_1(u,v) = \frac{1}{n}\indi_{u=v},
\end{equation} 
and $\mathcal{P}_2$ be the uniform distribution on $[n]\times[n]$, i.e., 
\begin{equation}\label{eq:dist_p2}
p_2(u,v) = \frac{1}{n^2}
\end{equation} 
for all $(u,v)\in[n]\times[n]$. Then, we know that
$$
\overline{M}^2(\vecw,\vecw^\prime) = \E_{(i,j)\sim\mathcal{P}_1}[\innerps{\gradf_i(\vecw) - \gradf_i(\vecw^\prime)}{\gradf_j(\vecw) - \gradf_j(\vecw^\prime)}],
$$
and
$$
\overline{G}(\vecw,\vecw^\prime) = \E_{(i,j)\sim\mathcal{P}_2}[\innerps{\gradf_i(\vecw) - \gradf_i(\vecw^\prime)}{\gradf_j(\vecw) - \gradf_j(\vecw^\prime)}].
$$
Then we find the joint distribution $\mathcal{P}_3$ of $(s_i,s_j)$ where $kB \le i,j \le (k+1)B-h-1$ and $i\neq j$.
Since $\vecz_{s_t} = \tvecz_{s_t}$, we know that $s_t \neq I$ for all $t=kB, \ldots, (k+1)B-h-1$. Then conditioned on $I$, $(s_i, s_j)$ is uniformly distributed in $([n]\setminus\{I\}) \times ([n]\setminus\{I\})$. For any $u\in [n]$, we have
\begin{align*}
p_3(u,u) &= \probs{s_i=u,s_j=u} = \frac{1}{n}\sum_{\ell=1}^n\probs{s_i=u,s_j=u \mid I=\ell} \\
&= \frac{1}{n} \sum_{\ell=u} \probs{s_i=u,s_j=u \mid I=\ell} = \frac{1}{n(n-1)}.
\end{align*}
For any $(u,v)\in[n]\times[n]$ such that $u\neq v$, we have
\begin{align*}
p_3(u,v) &= \probs{s_i=u,s_j=v} = \frac{1}{n}\sum_{\ell=1}^n\probs{s_i=u,s_j=v \mid I=\ell}\\
&= \frac{1}{n}\sum_{\ell\neq u,v} \probs{s_i=u,s_j=v \mid I=\ell} \\
&= \frac{n-2}{n(n-1)^2}.
\end{align*}
Then, we know that
$$
p_3(u,v) = \frac{1}{(n-1)^2}p_1(u,v) + \frac{n(n-2)}{(n-1)^2}p_2(u,v).
$$
Therefore, for any $i,j$ such that $kB \le i,j \le (k+1)B-h-1$ and $i\neq j$, we have
\begin{align*}
& \E_{I,A\mid H_{k+1}, A_{kB}, \Gamma}[\innerps{ \gradf_{s_i}(\vecwb_{kB}) - \gradf_{s_i}(\tvecwb_{kB}) }{  \gradf_{s_j}(\vecwb_{kB}) - \gradf_{s_j}(\tvecwb_{kB})  }] \\
=& \E_{(s_i,s_j)\sim\mathcal{P}_3}[\innerps{ \gradf_{s_i}(\vecwb_{kB}) - \gradf_{s_i}(\tvecwb_{kB}) }{  \gradf_{s_j}(\vecwb_{kB}) - \gradf_{s_j}(\tvecwb_{kB})  }] \\
=& \frac{1}{(n-1)^2}\overline{M}^2(\vecwb_{kB}, \tvecwb_{kB}) + \frac{n(n-2)}{(n-1)^2} \overline{G}(\vecwb_{kB}, \tvecwb_{kB}).
\end{align*}

\subsection{Examples of Differential Gradient Diversity and Diversity-inducing Mechanisms}\label{sec:example-dgd}
\paragraph*{Generalized Linear Functions} We can show that for generalized linear functions, the lower bound in Theorem~\ref{thm:lbbs} still holds, \ie, for any
$\vecw,\vecw'\in\W$, $\vecw\neq\vecw'$, we have
$$
\overline{B}_\setS(\vecw,\vecw') \ge \frac{\min_{i=1,\ldots,n} \twonms{\vecx_i}^2 }{ \sigma_{\max}^2(\matX)}.
$$
To see this, one can simply replace $\gradf_i(\vecw)$ with $\gradf_i(\vecw)-\gradf_i(\vecw')$ in Appendix~\ref{prf:lbbs}, and define 
$a_i=\ell_i^\prime(\vecx_i\tsp\vecw) - \ell_i^\prime(\vecx_i\tsp\vecw')$. The same arguments in Appendix~\ref{prf:lbbs} still go through. Consequently, for i.i.d. $\sigma$-sub-Gaussian features, we have $ \overline{B}_\setS(\vecw,\vecw') \ge c_1d $ $\forall~\vecw,\vecw'\in\W$ with probability at least $1-c_2ne^{-c_3d}$; and for
Rademacher entries, we have $\overline{B}_\setS(\vecw,\vecw') \ge c_4d$ $\forall~\vecw,\vecw'\in\W$ with probability greater than $1-c_5e^{-c_6n}$.

\paragraph*{Sparse Conflicts} The result in Theorem~\ref{thm:sparse-conflict} still holds for $\overline{B}_\setS(\vecw,\vecw')$, \ie for all $\vecw,\vecw'\in\W$, $\overline{B}_\setS(\vecw,\vecw') \ge n/(\rho+1)$, where $\rho$ is the maximum degree of all the vertices in the conflict graph $G$. To see this, one should notice
that the support of $\gradf_i(\vecw)$ only depends on the data point, instead of the model parameter, and thus, in general, $\gradf_i(\vecw)$ 
and $\gradf_i(\vecw)-\gradf_i(\vecw')$ have the same support. Then, one can simply replace $\gradf_i(\vecw)$ with $\gradf_i(\vecw)-\gradf_i(\vecw')$ in
Appendix~\ref{prf:sparse-conflict} and the same arguments still go through.

\paragraph*{Dropout} When we analyze the stability of mini-batch SGD, we apply the \emph{same} algorithm to two different samples $\setS$ and $\setS^{(I)}$ that 
only differ at one data point. Since the algorithm is the same, the random dropout matrices $\mat{D}_1,\ldots,\mat{D}_n$ are also the same in the two instances.
Therefore, one can replace $\gradf_i(\vecw)$ with $\gradf_i(\vecw)-\gradf_i(\vecw')$, and the same arguments still work. 
Then, we know that when $\overline{B}_\setS(\vecw,\vecw') \le n$, we have $\overline{B}_\setS^{\sf{drop}}(\vecw,\vecw') \ge \overline{B}_\setS(\vecw,\vecw')$,
and when $\overline{B}_\setS(\vecw,\vecw') > n$, we have $\overline{B}_\setS^{\sf{drop}}(\vecw,\vecw') > n$.

\paragraph*{Stochastic Gradient Langevin Dynamics} For SGLD, we can make similar arguments as in dropout, since the additive noise vectors $\xi_1,\ldots,\xi_n$
are the same for the two instances. One can then show that when $\overline{B}_\setS(\vecw,\vecw') \le n$, we have $\overline{B}_\setS^{\sf{sgld}}(\vecw,\vecw') \ge \overline{B}_\setS(\vecw,\vecw')$, and when $\overline{B}_\setS(\vecw,\vecw') > n$, we have $\overline{B}_\setS^{\sf{sgld}}(\vecw,\vecw') > n$.